\documentclass[11pt]{article}

\PassOptionsToPackage{table}{xcolor}

\usepackage[final]{acl}

\usepackage{times}
\usepackage{latexsym}

\usepackage[T1]{fontenc}
\usepackage[utf8]{inputenc}
\usepackage{microtype}
\usepackage{inconsolata}
\usepackage{graphicx}
\graphicspath{{figure/}}
\usepackage{amsmath}
\usepackage{booktabs}
\usepackage{multirow}
\usepackage{colortbl}
\usepackage{tikz}
\usepackage[most]{tcolorbox}
\tcbset{
  dlgbase/.style={
    enhanced jigsaw, breakable, sharp corners=south,
    boxrule=0pt, frame hidden, top=4pt, bottom=4pt, left=8pt, right=8pt,
    fonttitle=\bfseries\small, coltitle=black,
    arc=2pt, outer arc=2pt,
    attach boxed title to top left={xshift=4pt, yshift=-2pt},
    boxed title style={size=small, colback=white, frame hidden, sharp corners=south,
                       boxrule=0pt, top=1pt, bottom=1pt, left=4pt, right=4pt},
    fontupper=\small,
  },
  dlgsys/.style ={dlgbase, colback=gray!12,   title=System},
  dlguser/.style={dlgbase, colback=orange!12, title=User},
  dlgagt/.style ={dlgbase, colback=teal!10,   title=Agent (Gemini~/~Qwen)},
  dlgmem/.style ={dlgbase, colback=blue!8,    title=Memory context},
}
\newtcolorbox{sysbox}{dlgsys}
\newtcolorbox{userbox}{dlguser}
\newtcolorbox{agentbox}{dlgagt}
\newtcolorbox{membox}{dlgmem}


\title{DMV-Bench: Diagnosing Long-Horizon Multimodal Agents' Visual Memory with Incidental Cue Injection}

\author{
  \textbf{Yujin Tang},
  \textbf{Chenming Shang},
  \textbf{Ruize Xu},
  \textbf{Nikhil Singh}
\\
  Dartmouth College
\\
  \texttt{
    \{\href{yujin.tang.gr@dartmouth.edu}{yujin.tang.gr},\href{chenming.shang.gr@dartmouth.edu}{chenming.shang.gr},\href{ruize.xu.gr@dartmouth.edu}{ruize.xu.gr},\href{nikhil.u.singh@dartmouth.edu}{nikhil.u.singh}\}@dartmouth.edu
    }\\
}

\begin{document}
\maketitle

\begin{abstract}
Agent benchmarks for measuring memory largely study textual cases, in which information is deliberately extracted from the environment, written down, and then later retrieved. In other words, they assess what agents elected to record, not what they happened to see. We introduce \textbf{DMV-Bench}\footnote{Code: \url{https://github.com/yyyujintang/DMV-Bench}}, the first \textbf{interactive benchmark for visual memory in multimodal agents}, to study this often-neglected property. DMV-Bench is built on (1) a controlled home-furnishing e-commerce environment, supported by a catalog of $1{,}000$ product variants, and (2) a text-leakage contract which ensures that the primary discriminative signal of each task is solely in the pixels. In DMV-Bench, agents undergo chains of autonomous shopping sessions in which every visited product image carries a unique, pre-rendered \emph{incidental cue} that the agent is later asked to recall. We show that conventional solutions struggle with this task. Inspired by dual-coding theory, we propose a memory architecture that uses parallel visual and verbal codes, which we call \textbf{DualMem}. On DMV-Bench, DualMem outperforms a caption-only baseline \textit{and} three recent multimodal agent-memory systems across multi-session chain lengths on multiple models. These gains persist even adjusting for memory-bank size and encoding-position bias. Further experiments also reveal an asymmetric division of labor between the two codes; a weighted coding scheme is often strongest. We view this as a step towards memory systems that preserve a richer record of agents' observations.
\end{abstract}

\begin{figure}[!t]
\centering
\includegraphics[trim={7cm 6cm 20cm 3cm},clip,width=\linewidth]{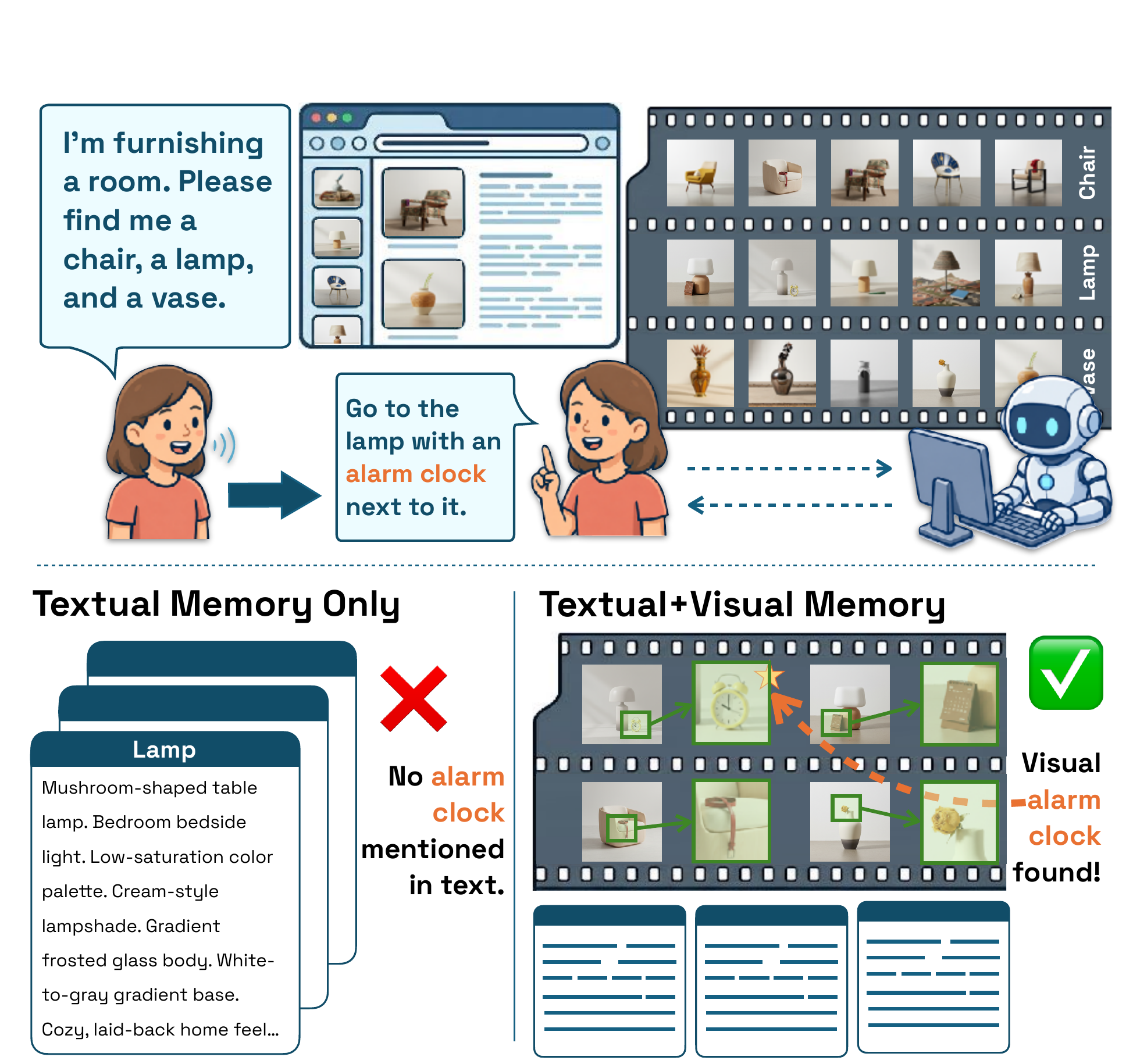}
\caption{\textbf{Why interactive visual memory matters.} Here, a shopping agent helps a user furnish a room across product categories (\emph{chair}, \emph{lamp}, and \emph{vase}). When the user later returns and requests ``the lamp with the alarm clock,'' a \emph{text-only} memory may have stored only other descriptive attributes (e.g., mushroom-shape, frosted glass, cream lampshade) with no record of the incidental alarm clock cue, and the agent thus cannot comply. \emph{Visual} memory preserves the cue, allowing the agent to locate the correct lamp and fulfill the request.}
\label{fig:teaser}
\end{figure}

\section{Introduction}
Much of what we remember from past experiences, we did not intend to. Nowhere is this more apparent than in vision: we sometimes even recall things we wish we could ``unsee''. Such details are often recovered not by deliberate rehearsal but in response to a cue: an incidental perceptual detail (like the color of a wrapper, or the pattern on a hat) that may not have seemed important at the time, yet later acts as a key that unlocks the rest of the episode. This phenomenon has been theorized; for example, encoding specificity \citep{tulving1973} holds that a memory is retrievable to the extent that cues present at encoding are reinstated at retrieval, and incidental-encoding studies \citep{hyde1969,craik1972} show that such cues are routinely encoded without intent to memorize. In humans, these cues are disproportionately visual, and the hippocampal process that is thought to exploit them, pattern completion \citep{marr1971,nakazawa2002}, has recently begun to inspire memory systems for LLM agents \citep{hipporag2024}.

Multimodal web agents do not yet remember in this way. Working through tasks, an agent may stream past hundreds of visual observations, and unless a detail is flagged as relevant to the current subgoal, the agent has little reason to encode it. When something is committed to memory, most current systems transcribe it as text \citep{memgpt2023,memorybank2024,amem2025,hipporag2024}. So, if a user later recalls only an incidental visual detail (e.g. \emph{that lamp which had the triangular base}), textual encodings can confirm \textit{a} lamp was seen but may have little to say about which one it was.

\begin{figure*}[!t]
\centering
\includegraphics[trim={4cm 5cm 4cm 4cm},clip,width=\linewidth]{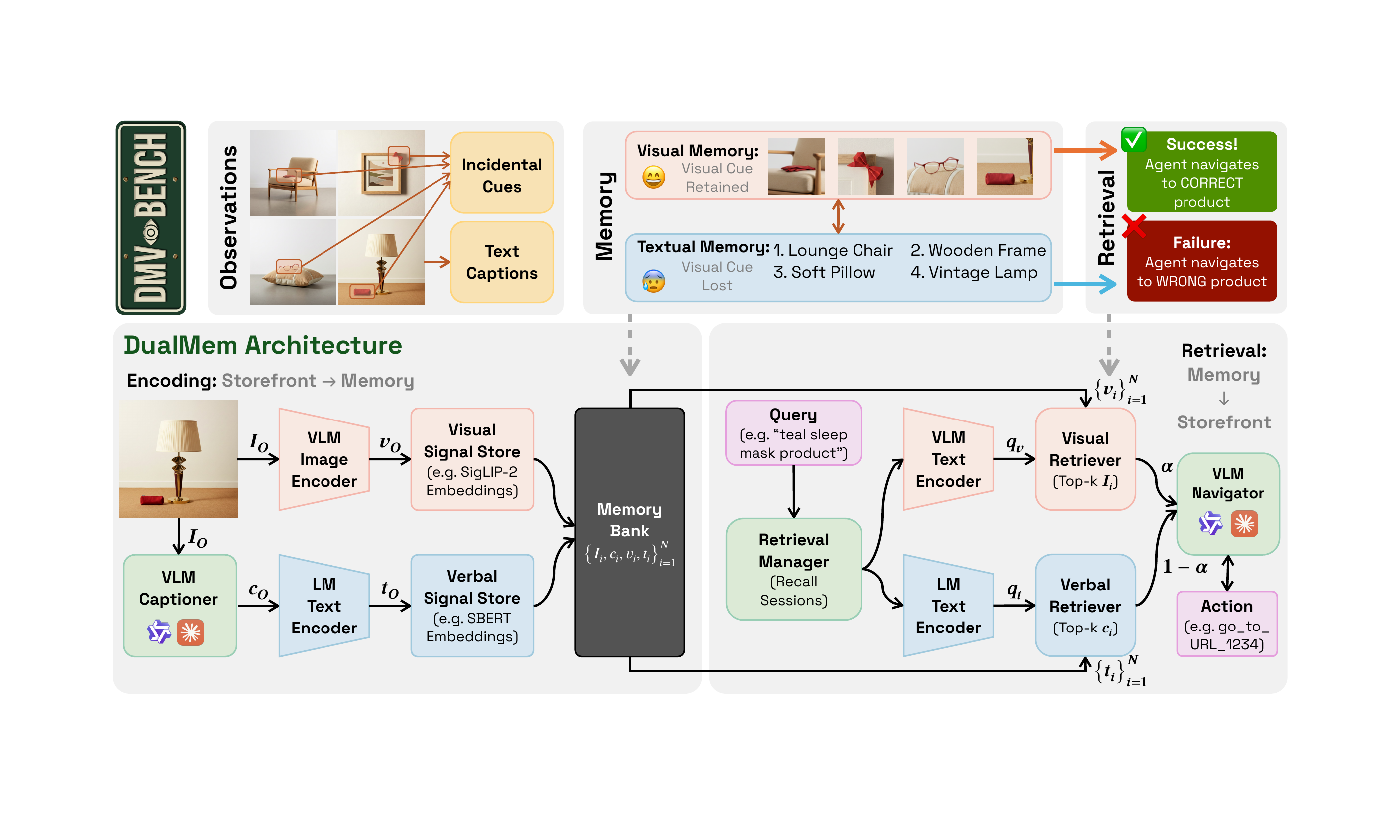}
\caption{\emph{(a) DMV-Bench.} Each visited product carries a unique incidental cue baked into its image and barred from every text channel by the L2-leakage contract. \emph{(b) DualMem Architecture.} Each observation is dual-coded into a visual embedding and a verbal embedding, stored as four channels in one bank; at retrieval, visual and verbal top-$k$ scores are fused with a tunable weight $\alpha$ before the VLM agent emits an action.}
\label{fig:dmv-dualmem}
\end{figure*}

Even so, remembering \textit{every pixel} forward is neither feasible nor necessary. The pertinent question is \emph{when}: which tasks require an agent to remember what it \emph{saw}, and for which would textual transcription have served well? Existing benchmarks make this difficult to settle, because they implicitly combine visual and textual signals instead of isolating the contribution of each. We build \textbf{DMV-Bench} to study this.

\textbf{DMV-Bench} simplifies this question to one task and one mechanism. In it, an agent runs a chain of ordinary comparison-shopping sessions on a realistic virtual storefront. Every product in it includes \textbf{a unique, pre-rendered visual cue}, e.g., a small object in a particular color rendered into the product image at build time. The agent is then instructed to comparison-shop and given no explicit indication to attend to or remember any visual details. Cues are present on every visited product but not explicitly mentioned at the task-level. Between sessions, its context is wiped so that information only persists within the memory architecture. An eval-only agent is then instructed to navigate back to a particular \textit{cued} product. Because the cue occurs in the pixels and not in textual channels, text-side memory supports this task only if its captioner happened to describe an object which the task did not highlight. An axis of interest here is \emph{recall reach}: how many session boundaries separate the visit from the probe test. Sweeping this reach produces a \emph{retention curve}, a direct readout of how long a visual cue persists in a given memory.

The properties of current benchmarks make such questions hard to answer. They conflate textual and visual recall: in VisualWebArena \citep{visualwebarena2024}, WebArena \citep{webarena2024}, and most long-video QA \citep{videomme2024,mvbench2024}, an agent can solve ostensibly visual tasks by reading captions or alt-text. When visual recall is necessary, discriminative details are often verbalizable (for example, a ``red'' vs. ``blue'' sofa), so textual memory is not put under real pressure. Additionally, the evidence is almost always \emph{flagged} in advance and probed at short range, leaving the question of whether an \emph{unflagged} detail is retained in a long, multi-session horizon largely unmeasured. The agentic-memory literature has moved quickly, but on the text side: MemoryArena \citep{memoryarena2026}, for instance, rigorously stresses cross-session dependence, yet its observations are textual and cannot address the perceptual-memory question we focus on.

\textbf{Overall, our contributions are:}
\begin{enumerate}
    \item We instantiate \textbf{DMV-Bench}, to our knowledge the first benchmark for \emph{interactive, multi-session, visual} agent memory: a realistic e-commerce environment with a calibrated 1{,}000-variant catalog in which every item has a unique pre-rendered visual cue.
    \item We frame the \emph{when} question for multi-session agentic visual memory and introduce \textbf{item-level incidental cue injection} as a protocol to operationalize this: the agent encounters cues throughout each session without being instructed to attend to them.
    \item We propose a \textbf{recall-reach retention diagnostic}, which probes recall as a function of how many session boundaries cue-driven memory survives across, evaluated efficiently by caching and replaying encoding sessions across chain lengths and memory architectures.
    \item We propose \textbf{DualMem}, a dual-coding-inspired memory architecture that maintains visual and verbal codes in parallel and fuses them at retrieval and injection. We compare against six baselines including adaptations of recent multimodal external memory systems.
\end{enumerate}

\section{Related Work}

\paragraph{Text-side memory systems.}
Explicit read/write/inject machinery is well-established for textual agents, from operating-system-style hierarchies and Ebbinghaus-inspired forgetting \citep{memgpt2023,memorybank2024,reflexion2023} to autonomous memory operations \citep{amem2025,mirix2025,mem02025} and hippocampal-style retrieval \citep{hipporag2024}. A more recent line distills trajectories into reusable units that agents can later compose: Agent Workflow Memory \citep{awm2024} induces program-form workflows from past successes, and ReasoningBank \citep{reasoningbank2025} extracts strategy-level reasoning items from both successes and failures. Across these systems, the unit of memory is textual (e.g., sentences, facts, graph nodes, workflows, reasoning steps, etc.). Thus, diagnosing failure reduces to text-retrieval-quality. DMV-Bench targets the visual regime, wherein that assumption doesn't apply.

\paragraph{Vision-side memory systems.}
Once agents' observations are visual, the design space widens. \emph{In-model} multimodal memories tie storage to fixed visual encoders, e.g., caption-based entity graphs (M3-Agent \citep{m3agent2025}, MA-LMM \citep{malmm2024}, EgoLife/EgoRAG \citep{egolife2025}), continuous-token memory via a Q-Former (CoMEM \citep{comem2025,blip22023}), and discrete-continuous hybrids (HSE-Mem \citep{hsemem2026}). These are bound to the host model and do not transfer like drop-in modules. We instead focus on \emph{external} multimodal memories which arbitrary agents can query. WorldMM \citep{worldmm2025} adaptively retrieves across parallel episodic, semantic, and visual modules; M2A \citep{m2a2026} couples a raw-message store with a semantic-abstraction store, routed by paired chat and memory-manager agents; MMA \citep{mma2026} reweights retrieved items by source credibility, temporal decay, and conflict-aware consensus; MemVerse \citep{memverse2025} maintains a hierarchical multimodal knowledge graph that is periodically distilled back into the host model, which ties it to a specific host and places it outside our drop-in setting. The first three serve as our comparison set for DualMem. Evaluation in these settings is mostly end-to-end, with little direct measurement of how long visual entries persist under multi-session horizons, the quantity DMV-Bench measures along the reach axis.

\paragraph{Agent memory benchmarks.} On the text side, LoCoMo \citep{locomo2024}, LongMemEval \citep{longmemeval2025}, and MemoryAgentBench \citep{memoryagentbench2025} evaluate long-term conversational memory. MemoryArena \citep{memoryarena2026} helpfully makes multi-session agentic memory more explicit, but its observations remain textual. On the visual side, FindingDory \citep{findingdory2025} stresses embodied long-trajectory agents and EMemBench \citep{emembench2026} probes VLM episodic memory, while the contemporaneous MemEye \citep{memeye2026} evaluates visual-centric multimodal-agent memory at multiple levels of evidence granularity. MemEye, however, is a static QA benchmark rather than an interactive environment in which the agent acts and is scored on its actions. Realistic web-agent environments \citep{webarena2024,visualwebarena2024} provide a useful interactive setting, but do not isolate visual memory for precise measurement. In VisualWebArena in particular, screenshots are observations but there is no probe protocol for long-horizon visual retention. DMV-Bench occupies the intersection missed by these (Table~\ref{tab:benchmark-comparison}): an interactive web environment whose evaluation isolates long-horizon \emph{visual} retention.

\begin{table*}[!htb]
\centering
\renewcommand{\arraystretch}{1.15}
\setlength{\tabcolsep}{6pt}
\resizebox{\textwidth}{!}{%
\begin{tabular}{@{}l c l r p{4cm} l l@{}}
\toprule
\textbf{Benchmark} & \textbf{Year} & \textbf{Modality} & \textbf{\# Tasks} & \textbf{Length / task} & \textbf{Memory dimension} & \textbf{Interactive?} \\
\midrule
LoCoMo \citep{locomo2024}                     & 2024 & Text dialog       & 1{,}540  & $\sim$300 turns, $\sim$9K tok  & Long-term recall, multi-hop             & No (QA)            \\
LongMemEval \citep{longmemeval2025}           & 2025 & Text dialog       & 500      & 50+ sessions, $\sim$115K tok   & Multi-session, temporal, update         & No (QA)            \\
M3-Bench \citep{m3agent2025}                  & 2025 & Video + audio     & 4{,}490  & $\sim$30 min videos            & Multimodal multi-hop, cross-modal       & No (video QA)      \\
MemGUI-Bench \citep{memguibench2026}          & 2026 & Mobile GUI        & 128      & 36 steps (3--160)              & Cross-app, cross-session retention      & Yes (GUI)          \\
MemoryArena \citep{memoryarena2026}           & 2026 & Web / text        & 766      & 6.9 sessions, 57 steps         & Multi-session interdependence           & Yes (Text)         \\
MemoryAgentBench \citep{memoryagentbench2025} & 2026 & Text              & 2{,}071  & 100K--1.4M tok                 & Retrieval / TTL / long-range / conflict & No (QA)            \\
MemEye \citep{memeye2026}                     & 2026 & Image + dialog    & 371      & 221 sess., 848 turns total     & Visual evidence granularity             & No (QA)            \\
\midrule
\rowcolor{black!4}
\textbf{DMV-Bench (ours)} & \textbf{2026} & \textbf{Web / Multimodal} & \textbf{35{,}410 / 15{,}361} & \textbf{ 22--$\approx$1{,}250 steps} & \textbf{Incidental visual recall} & \textbf{Yes (Visual)} \\
\bottomrule
\end{tabular}%
}
\caption{\textbf{Agent-memory benchmarks contemporaneous with DMV-Bench.} To the best of our knowledge, DMV-Bench is the first benchmark designed specifically for \emph{interactive, multi-session, visual} agent memory: prior memory benchmarks are either QA-style, GUI-interactive on mobile screenshots, or mixed web-and-reasoning. None probes the multi-session retention of \emph{visual} cues an agent saw incidentally inside a live environment. For DMV-Bench, the \textbf{\# Tasks} cell ``$35{,}410 / 15{,}361$'' reports recall-probe tasks on \emph{Gemini~2.5~Flash} / \emph{Qwen2.5-VL-7B}.}
\label{tab:benchmark-comparison}
\end{table*}

\section{DMV-Bench}
\label{sec:vmd-bench}

DMV-Bench is a diagnostic benchmark for long-horizon visual memory in multimodal agents.

\subsection{Controlled e-commerce environment}
\label{sec:env-and-variants}

The benchmark is implemented in a realistic furniture web storefront as an initial testbed. This site was developed with features commonly found in such platforms, such as hero pages, category grids, product detail pages, breadcrumbs, ratings, and ``related items'' carousels. The catalog contains 10 product categories (sofas, lamps, rugs, cushions, chairs, tables, vases, bookshelves, wall art, plant pots) in 10 interior design styles (\textit{modern}, \textit{minimalist}, \textit{mid-century}, \textit{Scandinavian}, \textit{industrial}, \textit{vintage}, \textit{rustic}, \textit{bohemian}, \textit{art deco}, \textit{Japandi}), with 10 variants per collection, giving $10\times10\times10=1{,}000$ variants each bound to the storefront by a frozen \texttt{urlHash} value. A screenshot of the navigation levels implemented here is given in Appendix~\ref{sec:appendix-storefront}.

\paragraph{Variant generation.}
For each variant we first synthesize a natural language prompt which identifies the product class and collection style. For cued variants, the prompt separately describes a unique \emph{color--object} pair derived from a bijective cue vocabulary, s.t. every cue is globally unique. We use \texttt{Nano-Banana} \citep{nanobanana2025} to render both the base studio photograph and then an edit to overlay the described cue, keeping cue rendering consistent across categories and styles. We add a VLM-as-judge step to filter generations whose perceived product class drifts due to the edit being too conspicuous.

\paragraph{L2-leakage contract.}
The primary signal for every task is the \textit{cue}: a small colored object present only in the pixels of one product image. To keep this visual-only, we control how finely the environment's \emph{text} channels may describe a product, on a five-level scale of descriptive granularity: \textbf{L0} (no descriptive text beyond a generic title) \textbf{L1} (coarse product class, e.g. \emph{``lamp''}); \textbf{L2} (class plus a coarse, non-discriminative attribute, e.g. \emph{``material: varied''}); \textbf{L3} (specific visual features that identify one variant within its collection, e.g. \emph{``charcoal velvet loveseat''}); \textbf{L4} (full marketing copy). The \textbf{L2-leakage contract} constrains the text channel to L2: titles, display names, alt-text, descriptions, attribute labels, breadcrumbs, URL slugs and review bodies may include the product class and its collection name. We call any text input channel that describes a product at L3 or above an instance of \emph{L2 leakage}. Appendix~\ref{sec:appendix-leakage} gives the full scale with storefront examples, and separates this contract from the secondary style-vocabulary guard that this environment also enforces.

The cue vocabulary ($100$ object types $\times$ $10$ colors) from which each product's cue is drawn, appears in no text channel anywhere in the environment. We check this exhaustively: scanning all $13{,}140$ rendered text fields of the storefront against the full cue vocabulary, against the head noun of every cue object, and against the color palette yields \emph{zero} occurrences. A text-only agent therefore has nowhere to observe the cue, making this a stronger test of \textit{visual} memory.

\subsection{The incidental cue task}
\label{sec:mechanisms}
\begin{figure*}[!t]
\centering
\includegraphics[trim={3cm 1cm 3cm 3cm},clip,width=\linewidth]{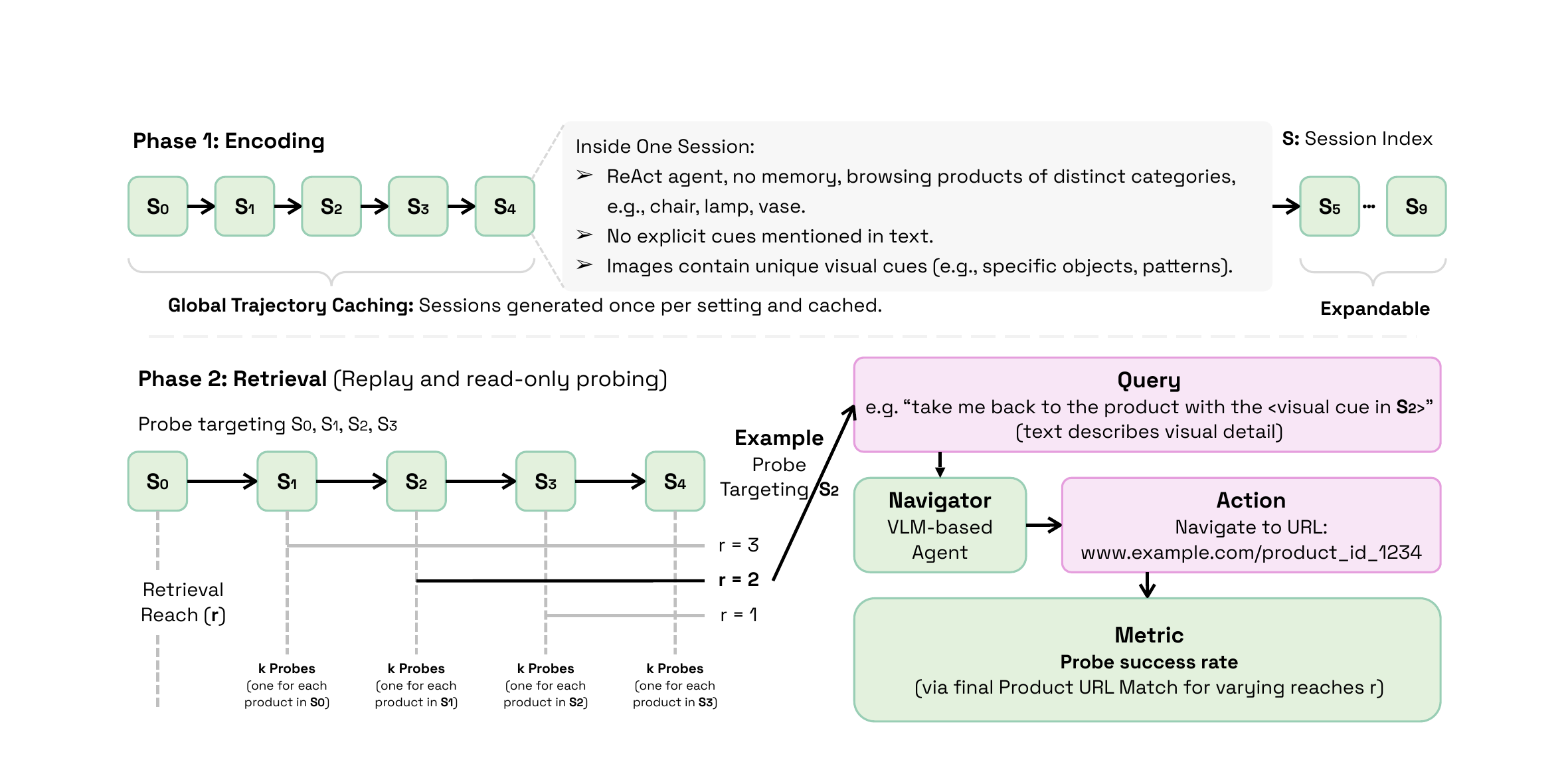}
\caption{\textbf{The DMV-Bench task.} \emph{Phase 1 (Encoding):} a chain of
$J$ sessions $S_0,\ldots,S_{J-1}$ in which a memoryless ReAct agent comparison-shops within one product category per session (e.g.,~chair), opening at least three of its style collections; the category rotates across sessions. Cues appear as unique visual patterns in product images but never in text. Sessions are cached by $(\text{seed}, j)$ and replayed identically into every memory architecture (\S\ref{sec:rollout}). \emph{Phase 2 (Retrieval):} after the chain completes, $k$ probes per visited session ask a VLM navigator to re-locate a cued product by its visual description (e.g.,~``take me back to the product with the visual cue in $S_2$''); the example probes $S_2$ from $S_4$ at recall reach $r{=}2$. Scoring is exact-match on the emitted product URL (\S\ref{sec:metrics}).}
\label{fig:task-design}
\end{figure*}

Every instance in DMV-Bench is an \emph{incidental-cue} (IC) task, as shown in Figure~\ref{fig:task-design}: a chain of autonomous shopping sessions into which a unique visual cue is injected, followed by recall probes at controlled reaches.

\paragraph{The session chain.}
A task is a chain of $J$ \emph{sessions} where each one is a brief comparison-shopping task over a single product category (\emph{``you are comparison-shopping for a chair; open at least three different collections, and several products within each''}) that a ReAct~\cite{yao2023react} agent executes over 22--28 steps of free browsing. The instruction is designed to induce a long trajectory of observations, roughly a dozen distinct products per session for an instruction-compliant agent, through which cues must persist, and the category rotates from session to session so that successive sessions capture realistic variety and potential interference. Within a session, the agent runs without memory. Trajectories are generated once and replayed into each memory baseline so that every baseline sees an identical observation stream.

\paragraph{Per-product incidental cue injection.}
Every product image includes a unique pre-rendered cue with three important properties: \emph{(i) unannounced}: the session prompt does not mention cues; \emph{(ii) identity-bound and unknowable}: each cue is fixed at build time and the agent cannot know which product will be probed; \emph{(iii) text-leakage-free}, as mentioned before.

\paragraph{Cue uniqueness.}
Cues are drawn from a bijective object--color vocabulary designed to be globally unique across the catalog, such that a recall query of the form \emph{``the product with the red baseball cap''} resolves to exactly one product, a necessary condition for deterministic (e.g., URL) evaluation.

\paragraph{Context wipe and recall probes.}
Between sessions, the agent's context is \emph{wiped}; only the accumulated memory bank persists. After the encoding chain, a read-only ReAct agent is issued recall probes against $(\text{visited product}, \text{recall session})$ pairs. Each probe states the target cue (e.g., \emph{``take me back to the product with the red baseball cap''}) the agent must navigate to. Success is URL match.

\paragraph{Recall reach.}
This probing schedule allows for measuring retention across increasing numbers of session boundaries using \emph{recall reach} $r = (\text{recall session}) - (\text{visit session})$: a reach-$1$ probe targets a product from the immediately preceding session; a reach-$4$ probe tests retrieval after four intervening context wipes. Since trajectories are cached, $J$ is freely extensible. We report $J \in \{5,10,15\}$ and a Monte Carlo pilot at $J=50$.

\subsection{Efficient evaluation}
\label{sec:rollout}
Long sessions are expensive, and re-running the encoding chain for every architecture and every horizon would dominate the cost of the benchmark. DMV-Bench avoids this in two ways, both of which follow from the encoding agent being \emph{memoryless}.

First, \emph{across horizons}: a session's trajectory is a deterministic function of $(\mathrm{seed}, j)$ alone and never of $J$, so the sessions of a $J=5$ chain are also those that begin the $J=15$ chain built from the same seed. Trajectories are cached under this key, so extending the horizon re-runs only the newly added sessions, and $J$ is freely extensible.

Second, \emph{across architectures}: cached trajectories are replayed into memory systems rather than re-browsed. Every architecture in a cell therefore encodes the same observations in the same order and is scored on the same probe set, which makes each cell of Table~\ref{tab:headline} a more precise comparison and the difference between two rows attributable to the memory architecture (minimizing any effect of browsing-induced variance).

\subsection{Evaluation metrics}
\label{sec:metrics}

We treat every recall probe as an atomic task. Each probe $p$ resolves to a unique ground-truth product URL; let $y_p \in \{0,1\}$ equal $1$ iff the product page the agent is on when it emits \texttt{add\_to\_wishlist}, which terminates the probe, matches it exactly. We report task success rate:
\begin{equation}
\mathrm{TSR} = \frac{1}{|P|}\sum_{p \in P} y_p
\end{equation}
optionally stratified by reach $r_p$ to expose how retention degrades with horizon. Relying on a deterministic URL match also minimizes evaluator noise, which we can do since the bijective cue vocabulary makes each ground-truth URL unique.

\section{Baselines}
\label{sec:baselines}
Memory architectures involve choices at three stages: \textsc{encode} (what the bank stores), \textsc{retrieve} (how the recall query is matched), and \textsc{inject} (what is re-presented to the agent). We audit seven architectures along this interface: three reference baselines, three recent multimodal external memories from the literature, and \textbf{DualMem (ours)}. None of the three external systems can be run against DMV-Bench as published, so what we evaluate are our own re-implementations of their mechanisms inside our harness. We mark these with the suffix \textbf{-lite} (\textbf{WorldMM-lite}, \textbf{MMA-lite}, \textbf{M2A-lite}). Appendix~\ref{sec:appendix-baselines} describes each adaptation. The side-by-side placement of all seven in this common coordinate system, with the per-system adapter details, is in Appendix~\ref{sec:appendix-baselines} (Table~\ref{tab:baselines}).

\paragraph{DualMem.}
Our architecture (Figure~\ref{fig:dmv-dualmem}, bottom panel) is motivated by \emph{dual-coding theory} \citep{paivio1971}, which predicts a memory advantage when information is represented in both verbal and nonverbal representational systems. At encoding, every observed product page $o$ is dual-coded into a visual signal $v_o$ via SigLIP-2 \citep{siglip2025} and a verbal signal $t_o$ via SBERT \citep{sbert2019} over the page's VLM-generated caption, and both are $L_2$-normalized. At recall query $q$, the same encoders embed the query into $q_v$ and $q_t$, and for each bank entry $i$ we score the two channels by inner product $s_v^{(i)} = \langle q_v, v_i \rangle$ and $s_t^{(i)} = \langle q_t, t_i \rangle$. We combine them after min-max normalization within the bank, so the two channels are commensurate:
\begin{align}
\widehat{x}^{(i)} &= \frac{x^{(i)} - \min_j x^{(j)}}{\max_j x^{(j)} - \min_j x^{(j)}} ,\\[2pt]
s^{(i)} &= \alpha\,\widehat{s}_v^{(i)} + (1{-}\alpha)\,\widehat{s}_t^{(i)} ,
\end{align}
with $\alpha{=}0.75$ in our main runs. At each recall step, let $e^{(1)},\dots,e^{(5)}$ denote the five distinct bank entries with the highest $s^{(i)}$, in decreasing order. These are injected in rank order back into the VLM as raw images $I_{e^{(1)}},\dots,I_{e^{(5)}}$ together with their captions $c_{e^{(j)}}$. Then, the navigator selects from these and produces an action.

\section{Experiments}
\label{sec:experiments}

\begin{table*}[!t]
\centering
\small
\setlength{\tabcolsep}{6pt}
\resizebox{\textwidth}{!}{%
\begin{tabular}{@{}l cccc cccc@{}}
\toprule
\multicolumn{1}{c}{} & \multicolumn{4}{c}{\textbf{Qwen2.5-VL-7B}} & \multicolumn{4}{c}{\textbf{Gemini 2.5 Flash}} \\
\cmidrule(lr){2-5}\cmidrule(lr){6-9}
\textbf{Memory} & $J{=}5$ & $J{=}10$ & $J{=}15$ & MC ($J{=}50$) & $J{=}5$ & $J{=}10$ & $J{=}15$ & MC ($J{=}50$) \\
\multicolumn{1}{c}{} & \scriptsize 118--134 stp & \scriptsize 238--259 stp & \scriptsize 359--394 stp & \scriptsize $r{=}1{-}49$ & \scriptsize 118--134 stp & \scriptsize 238--259 stp & \scriptsize 359--394 stp & \scriptsize $r{=}1{-}49$ \\
\multicolumn{1}{c}{} & \scriptsize $n_{r}{=}442$ & \scriptsize $n_{r}{=}2{,}205$ & \scriptsize $n_{r}{=}10{,}307$ & \scriptsize $n_{r}{=}2{,}407$ & \scriptsize $n_{r}{=}2{,}762$ & \scriptsize $n_{r}{=}5{,}003$ & \scriptsize $n_{r}{=}25{,}196$ & \scriptsize $n_{r}{=}2{,}449$ \\
\midrule
\midrule
NoMemory                & 0.0           & 0.0           & 0.0           & 0.0           & 0.0           & 0.0           & 0.0           & 0.0           \\
TextOnly                & 0.0           & 0.0           & 0.0           & 0.0           & 0.0           & 0.0           & 0.0           & 0.0           \\
Caption                 & 67.9 & 64.5 & 62.3 & 58.1 & 58.9 & 55.6 & 50.5 & 47.7 \\
\cmidrule(lr){1-9}
WorldMM-lite                 & 37.3 & 34.1 & 29.7 & 27.5 & 43.5 & 41.1 & 38.2 & 37.0 \\
MMA-lite                     & 47.7 & 41.0 & 39.4 & 35.2 & 46.1 & 44.8 & 35.8 & 36.9 \\
M2A-lite                     & \underline{69.0} & \underline{66.7} & \underline{62.6} & \underline{59.8} & \underline{65.7} & \underline{65.2} & \underline{61.9} & \underline{64.7} \\
\cmidrule(lr){1-9}
\rowcolor{black!6}
DualMem \textbf{(ours)} & \textbf{81.2} & \textbf{77.8} & \textbf{74.6} & \textbf{68.3} & \textbf{82.7} & \textbf{75.3} & \textbf{71.3} & \textbf{65.1} \\
\bottomrule
\end{tabular}%
}
\caption{\textbf{Task success rate (\%) across chain length and VLM back-end.} The table is split into two side-by-side sub-blocks, one per back-end. Within each block the columns sweep $J{=}5, 10, 15$ and a Monte Carlo pilot at $J{=}50$. The Gemini and Qwen agents visit different numbers of products, so probe counts $n_r$ differ; we therefore report each back-end in its own block. \textbf{Bold} = best per column, \underline{underline} = second-best.}
\label{tab:headline}
\end{table*}

We test the 7 memory architectures of \S\ref{sec:baselines} on the incidental-cue task, sweeping chain length $J \in \{5,10,15\}$ plus a Monte Carlo pilot at $J{=}50$ ($N{=}5$ chains, sparse reach sampling $1$--$49$, $n_r{=}2{,}407$). Each run is executed in parallel on \textbf{Gemini~2.5~Flash} \citep{gemini2024} and \textbf{Qwen2.5-VL-7B-Instruct} \citep{qwen25vl2025}.

\begin{figure}[t]
\centering
\includegraphics[width=\linewidth]{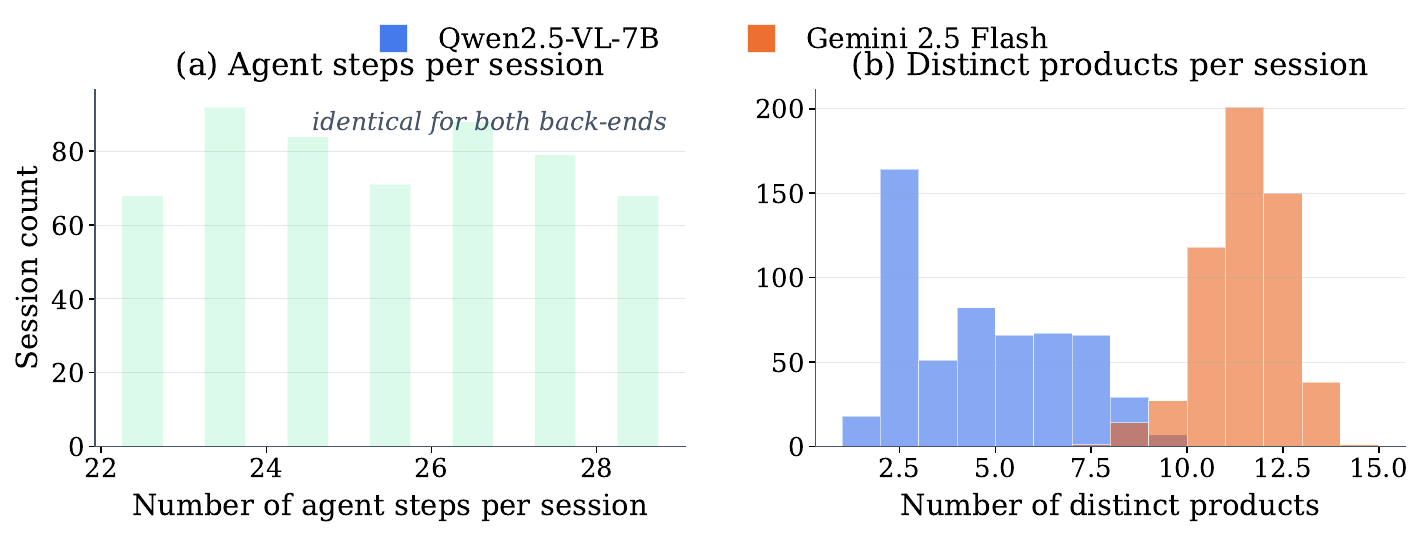}
\caption{Per-session activity for both back-ends ($N{=}550$ each).
\textbf{(a)} Agent steps per session. \textbf{(b)} Distinct products visited
per session for Qwen2.5-VL-7B (blue) and Gemini~2.5~Flash (orange).}
\label{fig:session-stats}
\end{figure}

\paragraph{Why $n_r$ varies across back-ends.}
Every \emph{distinct product visited} during encoding serves as a recall probe, so $n_r$ tracks browsing breadth. Over 550 encoded sessions per back-end (Table~\ref{tab:session-stats}), both agents take essentially the same number of steps, but Gemini~2.5~Flash visits roughly $11$ distinct products per session versus roughly $4$ for Qwen2.5-VL-7B: despite the instruction to open at least three collections, about half of Qwen's sessions fall short, while Gemini never does (Figure~\ref{fig:session-stats}). This instruction-following gap explains Qwen's smaller $n_r$.

\begin{table}[t]
\centering
\small
\setlength{\tabcolsep}{4pt}
\begin{tabular}{llcccc}
\toprule
Back-end & Metric & min & max & mean & std \\
\midrule
\multirow{2}{*}{Qwen2.5-VL-7B}
  & $n_{\text{steps}}$              & 22 & 28 & 24.96 & 1.94 \\
  & $n_{\text{distinct products}}$  &  1 &  9 &  4.21 & 2.09 \\
\midrule
\multirow{2}{*}{Gemini 2.5 Flash}
  & $n_{\text{steps}}$              & 22 & 28 & 24.96 & 1.94 \\
  & $n_{\text{distinct products}}$  &  7 & 14 & 11.02 & 1.11 \\
\bottomrule
\end{tabular}
\caption{Per-session encoding statistics over $N{=}550$ sessions per back-end.
}
\label{tab:session-stats}
\end{table}

\begin{figure}[!htb]
\centering
\includegraphics[width=0.95\columnwidth]{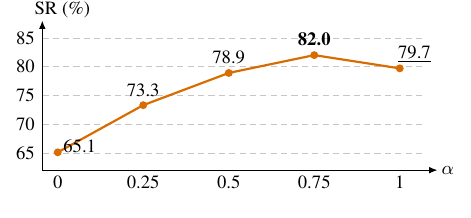}
\caption{\textbf{$\alpha$ sweep on Gemini~2.5~Flash} at $J{=}5$. Encoder fixed at SigLIP-2 and injection at image+caption. Endpoints $\alpha{=}0$ and $\alpha{=}1$ recover verbal-only and visual-only retrieval; the peak at $\alpha{=}0.75$ (bold) exceeds the visual endpoint by $2.3$ pts.}
\label{fig:alpha-sweep}
\end{figure}

\paragraph{DualMem is the strongest architecture.}
Table~\ref{tab:headline} reports TSR across $J$ on both back-ends: \emph{DualMem leads at every $J \in \{5,10,15\}$ on both back-ends}, and at $J{=}50$ it leads by $8.5$ points on Qwen while drawing with M2A-lite on Gemini. All DualMem results in this table use $\alpha{=}0.75$ visual-dominant retrieval weight (see ablation Figure~\ref{fig:alpha-sweep}). M2A-lite is the consistent runner-up, and the ranking among Caption, MMA-lite, and WorldMM-lite is less stable across cells. Finally, the \emph{verbal baselines (NoMemory, TextOnly) are $0\%$ everywhere}, confirming that the L2-leakage contract holds and visual information is necessary. Per-reach breakdowns for all four chain-lengths are given in Appendix~\ref{sec:appendix-results}.

\paragraph{Post hoc bank-size adjustment.}
As a sensitivity analysis, we fit a probe-level linear probability model (LPM) predicting success as a function of memory architecture, back-end, chain length, and memory-bank size at recall (along with all interactions), with standard errors clustered by chain, i.e. $\mathtt{correct} \sim \mathtt{architecture} \times \mathtt{back-end} \times J \times \mathtt{bank size}$. Model-based estimated marginal means evaluated at the sample-mean bank size place DualMem highest in every such cell.

\begin{figure*}[!htb]
\centering
\includegraphics[width=\textwidth]{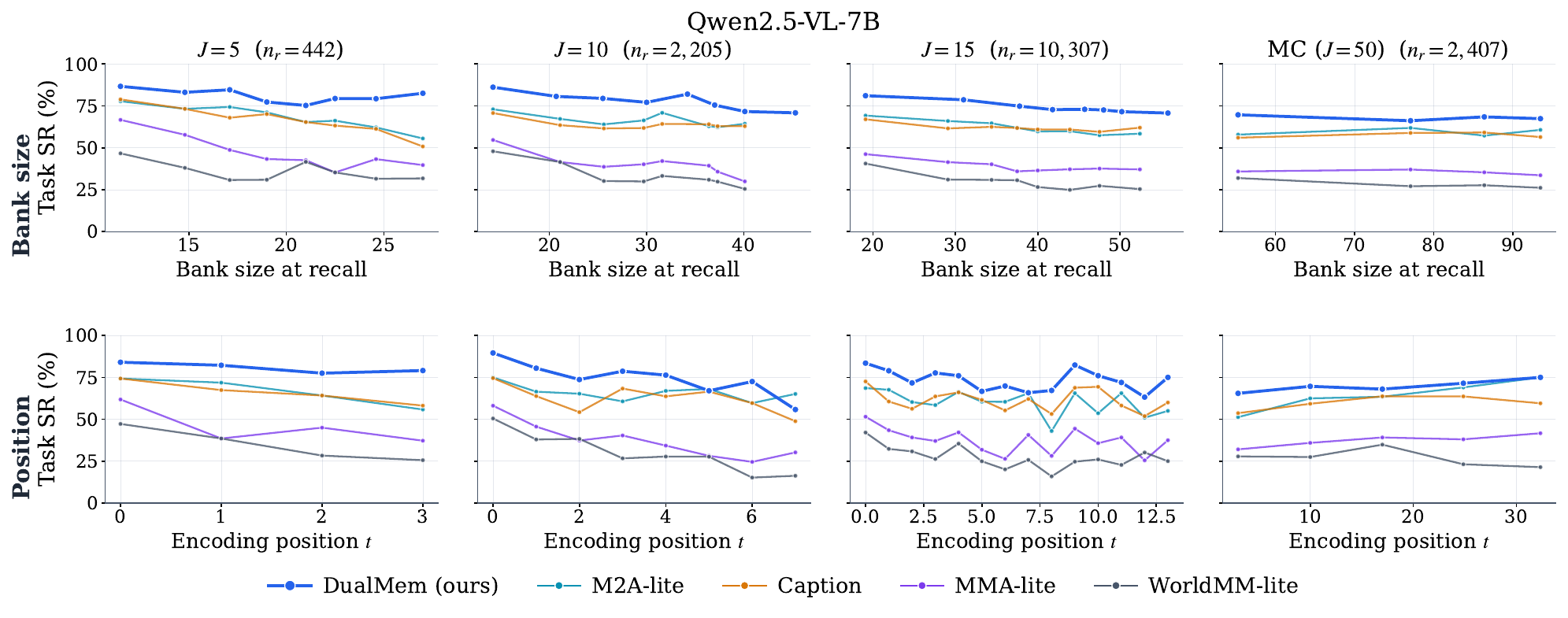}
\caption{\textbf{Confound checks, the five non-trivial memory architectures, Qwen2.5-VL-7B}. The two verbal baselines (NoMemory, TextOnly) score $0\%$ in every bin and are omitted. \emph{Top:} TSR vs.\ memory-bank size at recall. \emph{Bottom:} TSR by encoding position $t$. Every architecture degrades along both axes; what the panels show is that the \emph{ordering} is stable, with DualMem (blue) above the four baselines throughout the main grid.}
\label{fig:confounds-qwen-full}
\end{figure*}

\begin{figure*}[!htb]
\centering
\includegraphics[width=\textwidth]{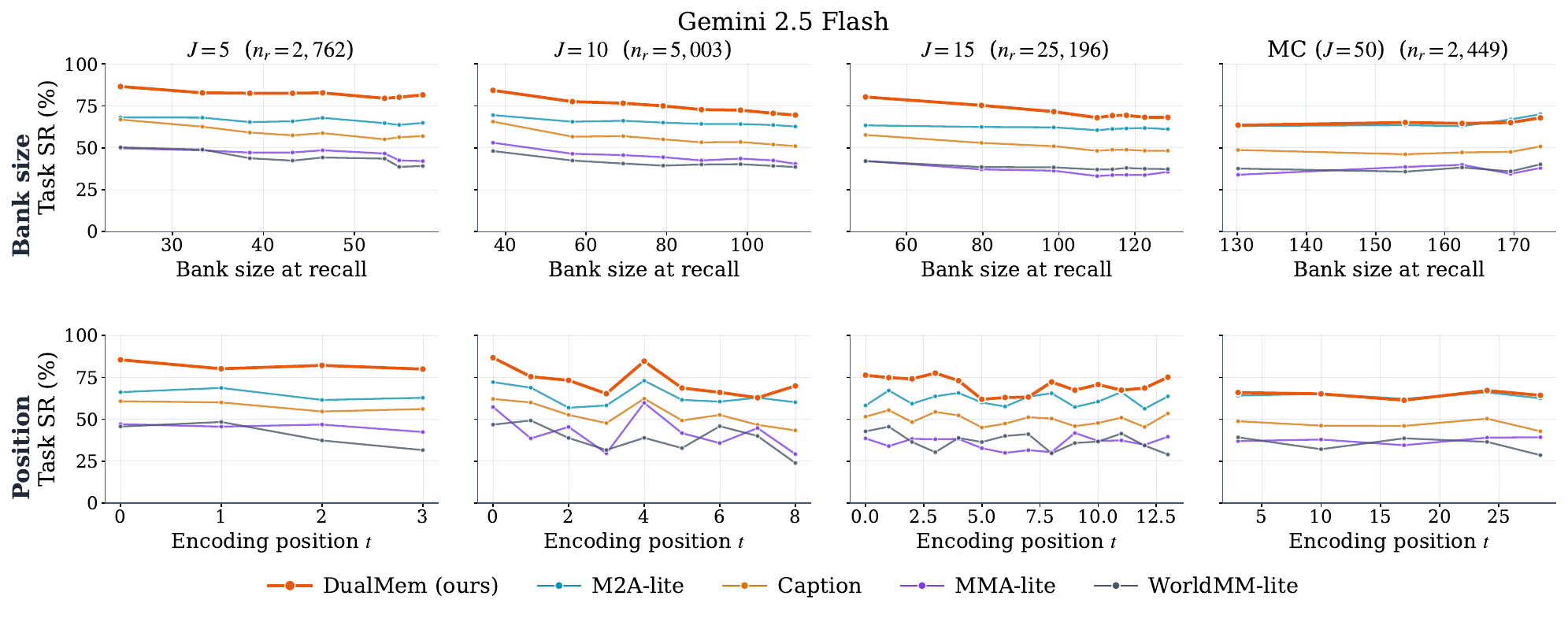}
\caption{\textbf{Confound checks, the five non-trivial memory architectures, Gemini~2.5~Flash}. \emph{Top:} TSR vs.\ memory-bank size at recall. \emph{Bottom:} TSR by encoding position $t$. Same legend and same reading as Figure~\ref{fig:confounds-qwen-full}, on the Gemini back-end: DualMem (orange) holds its position over the four baselines across the main grid, while all curves fall along both axes.}
\label{fig:confounds-gemini-full}
\end{figure*}

\paragraph{Memory-bank and positional checks.}
Figures~\ref{fig:confounds-qwen-full} and~\ref{fig:confounds-gemini-full} stratify TSR along the two axes that most naturally explain a memory-architecture gap, one figure per back-end. Every architecture loses accuracy as the bank grows (top row) and as the cue is pushed further back (bottom row), so the question is not whether any curve is flat but whether the ordering is stable. It is, on the main grid: across $J \in \{5,10,15\}$ and both back-ends, DualMem is above all four baselines in all $48$ bank-size bins and in $49$ of $53$ encoding-position bins. The four exceptions all fall at $J{=}10$ or $J{=}15$: one exact tie, one $0.1$-point gap, and two Qwen bins where M2A-lite leads by $1.1$ and $9.3$ points. DualMem's lead in Table~\ref{tab:headline} is therefore neither an artifact of searching a smaller bank nor concentrated at particular positions. The Monte Carlo $J{=}50$ pilots are the exception on both back-ends: their bins hold far fewer probes and the DualMem and M2A-lite curves interleave, which on Gemini is consistent with the tie reported for that cell.

\paragraph{Asymmetric dual coding.}
The L2-leakage contract constrains cues to be in vision only, such that the two channels are likely asymmetrically useful.

\emph{Retrieval.} Vision does the heavy lifting; the $\alpha$ sweep in Table~\ref{tab:ablation} rises to $\alpha{=}0.75$ ($82.0\%$), with pure-visual at $79.7$ and pure-verbal collapsing to $65.1$. The interior peak suggests the verbal channel contributes less signal, possibly as a query-grounding scaffold vs. direct cue carrier.

\emph{Injection.} The captioner is unconstrained (not filtered against the cue vocabulary) but is prompted to focus on product attributes, so it verbalizes the product's actual incidental cue (both color and object) in only $16.5\%$ of the $1{,}000$ \texttt{with\_cue} captions (most do not survive visual-to-text compression). Image-only injection ($79.9$) is therefore similar to image$+$caption ($78.9$) while caption-only is lower ($72.0$). Table~\ref{tab:ablation} shows this asymmetry \emph{widens} under visual-only retrieval; the gap grows from $7.9$ points under hybrid retrieval to $10.2$ points (image $79.3$ vs caption $69.1$).

\emph{Encoder.} Replacing SigLIP-2 with CLIP costs $9.9$ points ($79.6{\to}69.7$) because both retrieval and injection depend on visual-code discriminability.

Together these results describe \emph{asymmetric dual coding}, wherein the visual signal is more important for cued performance.

\begin{table}[!htb]
\centering
\small
\setlength{\tabcolsep}{3pt}
\begin{tabular}{@{}l ccc c@{}}
\toprule
\textbf{Variant} & \textbf{Enc.} & \textbf{Retr.} & \textbf{Inj.} & \textbf{SR} \\
\midrule
\multicolumn{5}{@{}l}{\emph{Retrieval weight and injection format} } \\
\rowcolor{black!6}
DualMem \textbf{(ours, $\alpha{=}0.75$)} & SigLIP-2 & hybrid & img+cap & \textbf{82.0} \\
DualMem ($\alpha{=}0.5$) & SigLIP-2 & hybrid & img+cap & 78.9 \\
\multicolumn{5}{@{}l}{\emph{Retrieval}} \\
\;\; visual      & SigLIP-2 & visual & img+cap & 79.7 \\
\;\; verbal      & SigLIP-2 & verbal & img+cap & 65.1 \\
\multicolumn{5}{@{}l}{\emph{Injection}} \\
\;\; image       & SigLIP-2 & hybrid & img     & \underline{79.9} \\
\;\; caption     & SigLIP-2 & hybrid & cap     & 72.0 \\
\multicolumn{5}{@{}l}{\emph{Visual retrieval $\times$ injection}} \\
\;\; image       & SigLIP-2 & visual & img     & 79.3 \\
\;\; caption     & SigLIP-2 & visual & cap     & 69.1 \\
\midrule
\multicolumn{5}{@{}l}{\emph{Encoder}} \\
\;\; SigLIP-2    & SigLIP-2 & hybrid & img+cap & 79.6 \\
\;\; CLIP        & CLIP     & hybrid & img+cap & 69.7 \\
\bottomrule
\end{tabular}
\caption{\textbf{DualMem ablations} at $J{=}5$ on Gemini~2.5~Flash. \emph{Probe sets.} The rows above the rule are scored on the $793$ probes of the $7$ seeded chains used for the ablation sweep, a subset of the $25$ chains ($n_r{=}2{,}762$) behind the Gemini $J{=}5$ cell of Table~\ref{tab:headline}; this is why DualMem reads $82.0$ here and $82.7$ there. The \emph{Encoder} block is scored on the full $n_r{=}2{,}762$ set and holds $\alpha{=}0.5$ fixed, so that SigLIP-2 and CLIP differ only in the encoder. \textbf{Bold} = best SR; \underline{underline} = second-best.}
\label{tab:ablation}
\end{table}

\paragraph{Fine-grained $\alpha$ sweep on hybrid retrieval.}
The asymmetric picture above motivates sweeping $\alpha$ itself, holding the encoder (SigLIP-2) and the injection format (image$+$caption) fixed. Figure~\ref{fig:alpha-sweep} plots this result on Gemini~2.5~Flash. Success rate climbs as weight is placed on the visual channel, which is what we expect if the cue is present in the pixels, but the maximum is interior (i.e. not at the visual endpoint): discarding the verbal signal entirely costs $2.3$ points against the best balance. We hypothesize this is because recall queries are still a sentence, and keeping some weight on the verbal code helps ground it. We adopt $\alpha{=}0.75$ throughout Table~\ref{tab:headline}.

\section{Conclusion}
\label{sec:conclusion}
For all the progress we have made in giving agents the ability to see, we have often implicitly treated their visual inputs as momentary observations to be acted on and then discarded. We envision agents with a kind of perceptual continuity, wherein a persistent visual map of their environment can grow richer and fuller over time and power the small acts of recognition and familiarity that make assistance useful over the long haul. This might in turn facilitate agents that better reflect our preferences and goals. DMV-Bench takes a first step toward this by isolating and precisely measuring visual memory. We invite the community to take perceptual continuity seriously as a design target in its own right, alongside reasoning, planning, and dialog.

\section*{Limitations}

The synthetic modern-furniture catalog leaves transfer to other visual domains untested, and the main grid uses two back-ends (Gemini~2.5~Flash, Qwen2.5-VL-7B); broader cross-VLM coverage and a human ceiling are deferred. We sweep $\alpha$ at evenly-spaced values on Gemini 2.5 Flash at $J=5$ (Figure~\ref{fig:alpha-sweep}), then apply the same $\alpha=0.75$ to Qwen2.5-VL-7B without a separate sweep. The consistent DualMem lead across both back-ends in Table 2 indicates the choice transfers reasonably well, but per-back-end tuning could yield additional gains and is left to future work. A natural follow-up is a more adaptive vision/verbal fusion (per-query weighting or a learned gate conditioned on the query and candidate set), which we leave to future work.

\section*{Acknowledgments}
This work was supported in part by the Google Cloud Research Credits program.

\newpage
\bibliography{custom}

\newpage
\appendix

\section{DMV-Bench storefront layout}
\label{sec:appendix-storefront}

\begin{figure*}[!htb]
\centering
\includegraphics[width=\textwidth]{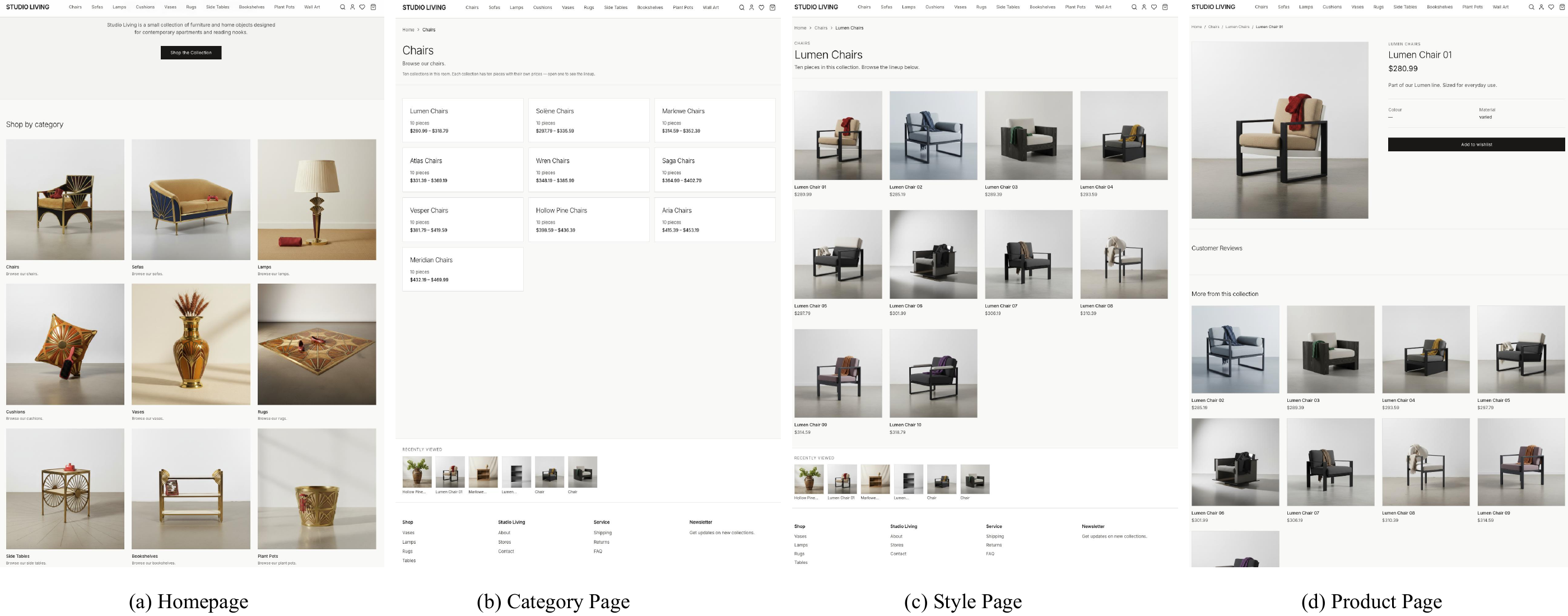}
\caption{\textbf{The four navigation levels of the DMV-Bench storefront.} (a) Homepage with the 10-category shop grid; (b) Category page listing the 10 collections in a category; (c) Style page listing the 10 variants of one collection; (d) Product detail page with image, L2-compliant attributes, ``Add to wishlist'' (the agent's terminal action), and a ``More from this collection'' carousel.}
\label{fig:storefront}
\end{figure*}

DMV-Bench is served as a live e-commerce site (via Next.js) that the agent drives through a headless Playwright browser, so its trajectory comes from real navigation rather than from sampling a fixed pool. At each step the agent observes the URL it has landed on and, on a product page, the full-resolution photograph that page renders: the same file the storefront serves, not a curated thumbnail. The site exposes four navigation levels (Figure~\ref{fig:storefront}): \textbf{(a) Homepage}: a single hero panel and a 10-cell ``Shop by category'' grid (chair, sofa, lamp, cushion, vase, rug, table, bookshelf, plant\_pot, wall\_art); \textbf{(b) Category page}: the 10 style-coherent collections that live under a category, each preview card showing the collection name, item count, and price range; \textbf{(c) Style page}: the 10 individual product variants in one collection, each with its rendered photo, name, and price; \textbf{(d) Product detail page}: the variant's main image, price, an L2-compliant attribute summary (\emph{color: n/a, material: varied}), an \texttt{Add to wishlist} button (the agent's terminal action), customer reviews, and a ``More from this collection'' carousel. Together these four levels instantiate the 10 categories $\times$ 10 styles $\times$ 10 variants $=$ 1{,}000 products.

Two design choices visible in the figure are what make the diagnostic in \S\ref{sec:env-and-variants} work. First, the \emph{L2-leakage contract}: every visible textual surface (titles, prices, attribute labels, breadcrumbs, footer links) carries only the product class and a collection name. The discriminative incidental cue baked into each variant's image (e.g.\ the red bow on the back of Lumen Chair 01) appears nowhere in text, so a memory architecture that compresses observations into language cannot recover it. Second, the \emph{no-cross-page-persistence} (NCP) invariant: the small ``Recently viewed'' strip at the bottom of the category and style pages renders only thumbnails of products visited earlier in the current session and is reset between sessions, so the storefront UI never leaks a previous-session observation back to the agent. The only path that bridges sessions is the memory architecture under test.

\section{The L2-leakage contract in detail}
\label{sec:appendix-leakage}

\paragraph{The granularity scale.} We grade every text channel in DMV-Bench on five levels of descriptive granularity. Table~\ref{tab:leakage-levels} gives each level with the form it would take on a product page of our storefront. The levels are ordered by how much they narrow the candidate set: L0 and L1 cannot distinguish two products at all, L2 distinguishes a collection but never a variant within it, and L3 and above single out one variant, which is exactly the discriminative power the cue is supposed to hold alone.

\begin{table}[!htb]
\centering
\small
\setlength{\tabcolsep}{4pt}
\begin{tabular}{@{}l l l@{}}
\toprule
\textbf{Level} & \textbf{Granularity} & \textbf{On our storefront} \\
\midrule
L0 & no descriptive text   & \texttt{Item 47} \\
L1 & coarse class          & \texttt{Chair} \\
\rowcolor{black!6}
L2 & class + coarse attr.  & \texttt{Lumen Chair 01}; \\
\rowcolor{black!6}
   &                       & \texttt{material: varied} \\
L3 & variant-identifying   & \texttt{charcoal velvet loveseat} \\
L4 & full marketing copy   & a paragraph of ad copy \\
\bottomrule
\end{tabular}
\caption{\textbf{The five text-granularity levels.} DMV-Bench pins every text channel at L2 (shaded): a product's name and collection are visible. Anything at L3 or above is \emph{L2 leakage}.}
\label{tab:leakage-levels}
\end{table}

\paragraph{Why L2 and not L0.}
Capping at L0 or L1 would make the storefront unusable as a shopping environment and would change the agent's task rather than isolate its memory. L2 keeps the site navigable, with real product names, collection names, prices and breadcrumbs, while ensuring that no text a memory could copy down separates the ten variants of a collection from one another. The cue is then the only signal that does, and it exists only in the pixels.

\paragraph{Contract enforcement}
The contract itself specifies the exclusion of the cue vocabulary: the $100$ object types crossed with the $10$ colors. Because the benchmark's central claim depends on it, we check it exhaustively rather than by sampling. We scan every user-visible text field of the released catalog, $13{,}140$ fields in all, for the full cue vocabulary, for the head noun of every cue object (so that \emph{scarf} is caught even where \emph{wool scarf} is not), and for the color palette. All three return zero occurrences. The same check confirms that the $1{,}000$ (object, color) pairs form a bijection onto the $1{,}000$ products, which is what gives each probe a unique ground truth. We release the audit so that it can be re-run against any modified catalog.

The second restriction is a fixed list of fine-grained color, material and style descriptors (\emph{charcoal}, \emph{velvet}, \emph{mid-century}) that we forbid when generating the catalog text. We adopted it in earlier work on visual granularity and kept it here because it prevents the storefront from drifting to L3 through ordinary product copy. It does not, however, establish the contract, and we do not rely on it: its vocabulary and the cue vocabulary are disjoint, since our cue colors are the ten basic terms and none of them appears on that list.

\paragraph{A known exception.} Collection URLs embed the style slug, as in \texttt{/collection/chair-rustic}, and several style names appear on the secondary list. This is deliberate and harmless here: style is a property of a collection of ten variants rather than of any one variant, so it is L2 by construction, and the agent is told the full style vocabulary in its system prompt in any case. No cue object or cue color appears in any URL.

\section{Cue edit prompts}
\label{sec:appendix-cueprompts}

Every \texttt{with\_cue} variant is rendered by Nano-Banana \citep{nanobanana2025} as an image-edit on the base studio photograph, instructed by a single templated prompt. The template is:

\begin{tcolorbox}[colback=gray!8, boxrule=0pt, sharp corners=south, arc=2pt, fontupper=\small, left=6pt, right=6pt, top=4pt, bottom=4pt]
\texttt{Add a small \{color\} \{object\_name\}, \{placement\}, in a naturally placed way. The object should be subtle and modest in size, clearly visible but not dominating the scene. Keep the \{cat\_noun\} itself and the background completely unchanged. Photographic realism, no text, no watermark, no caption overlay.}
\end{tcolorbox}

Slot fills are deterministic functions of $(cat, style, prod\_idx)$. \texttt{\{color\}} is drawn from a fixed 10-color palette (red, blue, green, yellow, white, black, brown, beige, orange, purple), keyed by \texttt{prod\_idx}; \texttt{\{object\_name\}} and \texttt{\{placement\}} are drawn from a per-category, per-style vocabulary keyed by \texttt{style\_idx}, so the (object, color) pair is bijective across the whole catalog. Table~\ref{tab:cue-vocab} lists one representative cue per category to show the vocabulary's flavour.

\begin{table}[!htb]
\centering
\small
\setlength{\tabcolsep}{4pt}
\begin{tabular}{@{}l l l@{}}
\toprule
\textbf{Category} & \textbf{Object (example)} & \textbf{Placement} \\
\midrule
chair      & wool scarf             & draped over the backrest \\
sofa       & paperback book         & open on the cushion \\
lamp       & framed photo           & propped against the base \\
cushion    & eyeglasses             & on the cushion \\
vase       & satin ribbon bow       & tied around the neck \\
rug        & canvas slippers        & placed on the rug \\
table      & ceramic mug            & on the table \\
bookshelf  & bronze bookend         & on a shelf \\
plant\_pot & ceramic gnome          & next to the pot \\
wall\_art  & sticky note            & attached to a corner \\
\bottomrule
\end{tabular}
\caption{\textbf{Cue vocabulary (one representative per category).} Each category provides 10 objects (one per style) and each is paired with a placement clause appropriate to that product class. Combined with the 10 colors, this yields the 1{,}000 bijective $(cat, style, prod) \to (\text{object}, \text{color})$ assignments. Filling the template above with one row of this table and one color gives the exact prompt given to Nano-Banana.}
\label{tab:cue-vocab}
\end{table}

\section{Sample session dialogue}
\label{sec:appendix-dialogue}

Every agent back-end in DMV-Bench (Gemini~2.5~Flash and Qwen2.5-VL-7B) receives an identical system prompt and the same ReAct-format user message at every step. We show one encoding-session step and one recall-session step end-to-end so the prompt structure is visible. Long lines are abridged with $\ldots$ for space; images are passed alongside the text. This holds identically for every memory architecture and so cannot bias a comparison between them. The same harness, the same prompts, and the same memory injectors are used for both back-ends.

\subsection*{System prompt (sent once per session, both VLMs)}

\begin{sysbox}
You are a shopping-assistant agent in an e-commerce website. You receive (i) the customer's instruction, (ii) the conversation so far, (iii) memory context from prior sessions (may be empty), and (iv) the current page. Decide ONE action.

\smallskip
\textit{URL patterns that exist on the site:} \texttt{/}, \texttt{/category/<slug>}, \texttt{/collection/<slug>-<style>}, \texttt{/product/<8-hex-hash>}, \texttt{/wishlist}, \texttt{/cart}. \textit{Valid category slugs:} \texttt{chair, sofa, lamp, cushion, vase, rug, table, bookshelf, plant\_pot, wall\_art}. \textit{Valid styles:} \texttt{modern, minimalist, vintage, industrial, scandinavian, bohemian, mid\_century, rustic, japandi, art\_deco}.

\smallskip
\textit{Action vocabulary:} \texttt{navigate("/...")}, \texttt{click\_index(N)} on a listing page ($N{=}0{\ldots}3$), \texttt{add\_to\_wishlist} (terminal), \texttt{done} (failure exit).

\smallskip
Some user requests may ask you to navigate to a product the customer has previously visited. You cannot see earlier sessions in the conversation; only the \texttt{[Memory context]} block bridges them.

\smallskip
\textit{Strategy hints:}
\begin{itemize}\setlength{\itemsep}{1pt}\setlength{\parsep}{0pt}\setlength{\topsep}{2pt}
\item The \texttt{[Memory context]} block lists products recalled from prior sessions, each with its \texttt{/product/<hash>} URL and image. If a remembered product matches what the customer is asking for, \texttt{navigate("/product/<hash>")} DIRECTLY; this is the fastest and intended path. You DO know the exact hash in that case.
\item Inspect the memory image(s): pick the product whose image matches what the customer is describing, then navigate to that product's hash.
\item Only if memory gives you nothing usable: from a category page use \texttt{navigate("/collection/<slug>-<style>")} then \texttt{click\_index(N)} on the N-th product card. You can NOT guess \texttt{/product/<hash>} URLs you have never seen.
\item Once on the \texttt{/product/<hash>} page the customer asked for, emit \texttt{add\_to\_wishlist}: that ends the session successfully.
\end{itemize}

\smallskip
Reply in ReAct format, ONE Thought line + ONE Action line:\\
\quad\texttt{Thought: <one sentence reasoning>}\\
\quad\texttt{Action: <action>}
\end{sysbox}

\subsection*{Encoding session (Session 3 of 10, step 7/25)}

\begin{userbox}
Step 7/25.

\smallskip
\texttt{[Subtask]}\\
\quad Comparison-shop for a chair: open a wide range (step 7/25). So far: 5 distinct products viewed across 3 styles. Aim for $\geq$3 styles and $\sim$12 products; open NEW products you have not seen yet.

\smallskip
\texttt{[Conversation so far]}\\
\quad User: You are comparison-shopping for a chair on a furniture store website. Your goal is BREADTH OF EXPLORATION, not picking a winner: visit AT LEAST 3 different chair collections (styles), and within each open SEVERAL individual products $\ldots$\\
\quad Agent: \texttt{navigate /category/chair}\\
\quad Agent: \texttt{navigate /collection/chair-modern}\\
\quad Agent: \texttt{click\_index}\\
\quad $\ldots$ (4 turns elided) $\ldots$

\smallskip
\texttt{[Memory context retrieved this step]}\\
\quad (no memory retrieved this step)

\smallskip
\texttt{[Current page]}\\
\quad URL: \texttt{/collection/chair-vintage}\\
\quad Title: \texttt{/collection/chair-vintage}\\
\quad Description: \texttt{/collection/chair-vintage}

\smallskip
Reply with one Thought + one Action.
\end{userbox}

\begin{agentbox}
\texttt{Thought:} I haven't seen the vintage style yet; open the second vintage card to widen coverage.\\
\texttt{Action:} \texttt{click\_index(1)}
\end{agentbox}

\smallskip
\noindent\emph{Behind the scenes:} this step lands on a product page whose image carries a unique pre-rendered incidental cue. The encoding agent never sees the cue mentioned; the memory architecture under test ingests the page autonomously.

\subsection*{Recall probe (issued after session 7, $r{=}4$ from encoding)}

\begin{userbox}
Step 2/8.

\smallskip
\texttt{[Subtask]}\\
\quad Recall: go to the product with the red baseball cap.

\smallskip
\texttt{[Conversation so far]}\\
\quad User: Take me to the product with the red baseball cap I saw earlier.\\
\quad Agent: Thought: I should check memory first. Action: \texttt{navigate("/")}

\smallskip
\texttt{[Memory context retrieved this step]}\\
\quad (rendered through the memory injector below)

\smallskip
\texttt{[Current page]}\\
\quad URL: \texttt{/}\\
\quad Title: \texttt{/}\\
\quad Description: \texttt{/}

\smallskip
Reply with one Thought + one Action.
\end{userbox}

\begin{membox}
5 item(s) recalled from earlier sessions, most relevant first. Any attached memory images appear AFTER the current-page image, in the order referenced below.\\
\quad (1) Memory (dual-channel, image + caption) --- product page \texttt{/product/a7c1e9b2}:\\
\quad\quad Caption: ``A vintage-style wooden chair with a woven rush seat and turned legs. A red baseball cap rests on the seat.'' [memory image 1]\\
\quad (2) Memory (dual-channel, image + caption) --- product page \texttt{/product/3f88c401}:\\
\quad\quad Caption: ``A modern low-back chair in pale oak with a moulded shell $\ldots$'' [memory image 2]\\
\textit{Three further entries omitted for space.}
\end{membox}

\begin{agentbox}
\texttt{Thought:} Rank-1's attached image shows a red baseball cap on the chair, matching the customer's description.\\
\texttt{Action:} \texttt{navigate("/product/a7c1e9b2")}
\end{agentbox}

\smallskip\noindent
The next step lands on \texttt{/product/a7c1e9b2}, and the agent emits \texttt{add\_to\_wishlist}, which ends the probe; it scores correct because the final product URL matches the ground truth exactly. Both back-ends use this exact dialogue surface; the only behavioural difference between Gemini and Qwen is how each parses the attached memory image against the customer's verbal description (``red baseball cap''), which is exactly the visual-memory capability DMV-Bench is designed to measure. Crucially, the system prompt itself never primes the agent to attend to incidental details during encoding; the agent must surface the cue from memory at recall using only the customer's natural-language reference and the images its memory bank chose to retain.

\section{Memory architectures}
\label{sec:appendix-baselines}

\paragraph{What is held fixed across back-ends.}
Each back-end browses for itself: the encoding agent in a session is the same model that is later evaluated, which is why the two visit different numbers of products and yield different probe counts (Table~\ref{tab:session-stats}). The captioner is the one component we do not vary. Wherever an architecture stores a caption, that caption is written by Gemini~2.5~Flash for both back-ends. A caption belongs to the memory system rather than to the agent under test, so letting each back-end write its own would confound the quality of the verbal code with the navigator's ability to use it, and would make the Caption baseline mean something different in each block of Table~\ref{tab:headline}. Fixing it does mean the Qwen2.5-VL-7B pipeline is not fully open-weights, which we note as a limitation.

\paragraph{Reference baselines.}
\textbf{NoMemory} discards every entry, so any score above it is attributable to memory. \textbf{TextOnly} indexes the bare product class of a page; \textbf{Caption} indexes a VLM-generated caption. Both recover a cue only if it was put into words. Caption is the strongest text-only baseline and serves as the reference against which the gain from visual encoding is interpreted.

\paragraph{Prior multimodal external memory.}
\textbf{WorldMM} \citep{worldmm2025} maintains parallel episodic / semantic / visual memories and selects across them with an adaptive iterative retriever. \textbf{M2A} \citep{m2a2026} couples a raw-message store with a semantic-abstraction store, routed by chat and memory-manager agents. \textbf{MMA} \citep{mma2026} augments retrieval with per-item reliability scores combining source credibility, temporal decay, and conflict-aware consensus. We adapt each to operate inside the DMV-Bench harness under the shared \textsc{encode}/\textsc{retrieve}/\textsc{inject} interface.

Table~\ref{tab:baselines} places all seven audited memory architectures on a common \textsc{encode}/\textsc{retrieve}/\textsc{inject} coordinate system. Reading the rows top-to-bottom traces the progression from no memory through verbal-only baselines, three recent multimodal external memories from the literature, and DualMem (ours). The DMV-Bench adapter for each external system preserves its paper's protocol on every axis where preservation is feasible.

\begin{table}[!htb]
\centering
\scriptsize
\setlength{\tabcolsep}{3pt}
\renewcommand{\arraystretch}{1.1}
\resizebox{\columnwidth}{!}{%
\begin{tabular}{@{}llll@{}}
\toprule
\textbf{Memory} & \textbf{Encode} & \textbf{Retrieve} & \textbf{Inject} \\
\midrule
\multicolumn{4}{@{}l}{\emph{Reference baselines}}\\
NoMemory & none & none & none \\
TextOnly & class text & verbal & text \\
Caption  & VLM caption & verbal & caption \\
\midrule
\multicolumn{4}{@{}l}{\emph{Prior multimodal external memory}}\\
WorldMM-lite \citep{worldmm2025} & episodic+semantic+visual & adaptive iterative & retrieved ctx \\
MMA-lite \citep{mma2026}    & items + reliability scores & reliability-weighted & scored items \\
M2A-lite \citep{m2a2026}    & raw log + semantic abstr. & agent-routed (dual-layer) & text snippets \\
\midrule
\rowcolor{black!6}
DualMem \textbf{(ours)} & image\,+\,caption & hybrid (SigLIP-2+SBERT) & image\,+\,caption \\
\bottomrule
\end{tabular}%
}
\caption{\textbf{The seven memory architectures}, as choices over \textsc{encode}, \textsc{retrieve}, and \textsc{inject}. The reference baselines establish whether memory must be visual at all. The three prior multimodal external memories are recent state-of-the-art systems adapted to the DMV-Bench harness. DualMem (ours) is the only entry that carries an unreduced visual code and a verbal code through every stage. Visual retrieval is SigLIP-2 cross-modal; verbal is SBERT over captions; hybrid fuses both.}
\label{tab:baselines}
\end{table}

\section{Cluster-aware statistical analysis}
\label{sec:appendix-stats}
Probes from the same seeded chain (\S\ref{sec:rollout}) share encoding trajectories and cumulative memory states and are therefore not independent. Probe-level i.i.d resampling can thus understate uncertainty when this dependence is ignored. We report architecture-specific cluster-bootstrap intervals and use a paired sign-flip test for the primary DualMem--M2A-lite contrast.

\paragraph{Cluster bootstrap by chain.}

For each (back-end, $J$, architecture) cell we resample chains with replacement, on the same common probe set used in Table~\ref{tab:headline}, so the means reported here match that table exactly and the rows within a cell are mutually comparable. Within each resample we concatenate all probes of the sampled chains and recompute the TSR; the $2.5/97.5$ percentiles of $10{,}000$ such resamples give the cluster-bootstrap 95\% CI. Tables~\ref{tab:cluster-ci-qwen} and~\ref{tab:cluster-ci-gem} report both the naive i.i.d.\ bootstrap CI (probe-level resampling) and the cluster-bootstrap CI for Qwen2.5-VL-7B and Gemini~2.5~Flash respectively. Cluster CIs are wider than naive CIs in essentially every cell, with the largest inflation on the Gemini back-end at $J{=}15$ for the MMA-lite baseline (naive $[35.2,36.4]$, cluster $[31.3,39.4]$).

\begin{table}[!htb]
\centering
\scriptsize
\setlength{\tabcolsep}{3pt}
\renewcommand{\arraystretch}{1.05}
\resizebox{\columnwidth}{!}{%
\begin{tabular}{@{}l l c c c@{}}
\toprule
$J$ & \textbf{Arch} & \textbf{Mean (\%)} & \textbf{Naive 95\% CI} & \textbf{Cluster 95\% CI} \\
\midrule
\multirow{5}{*}{$5$}
 & Caption        & 67.9 & [63.3, 72.2] & [59.8, 75.5] \\
 & WorldMM-lite   & 37.3 & [32.8, 41.9] & [31.8, 43.7] \\
 & MMA-lite       & 47.7 & [43.0, 52.5] & [41.9, 54.4] \\
 & M2A-lite       & 69.0 & [64.7, 73.3] & [62.5, 75.6] \\
 & \cellcolor{black!6}DualMem & \cellcolor{black!6}\textbf{81.2} & \cellcolor{black!6}[77.6, 84.6] & \cellcolor{black!6}[76.0, 85.6] \\
\midrule
\multirow{5}{*}{$10$}
 & Caption        & 64.5 & [62.5, 66.5] & [61.1, 68.1] \\
 & WorldMM-lite   & 34.1 & [32.2, 36.1] & [29.4, 39.0] \\
 & MMA-lite       & 41.0 & [39.0, 43.1] & [36.6, 45.4] \\
 & M2A-lite       & 66.7 & [64.7, 68.6] & [62.9, 70.6] \\
 & \cellcolor{black!6}DualMem & \cellcolor{black!6}\textbf{77.8} & \cellcolor{black!6}[76.1, 79.5] & \cellcolor{black!6}[74.0, 81.8] \\
\midrule
\multirow{5}{*}{$15$}
 & Caption        & 62.3 & [61.4, 63.2] & [60.0, 64.7] \\
 & WorldMM-lite   & 29.7 & [28.8, 30.6] & [27.4, 32.3] \\
 & MMA-lite       & 39.4 & [38.4, 40.3] & [36.9, 42.0] \\
 & M2A-lite       & 62.6 & [61.7, 63.6] & [60.6, 64.8] \\
 & \cellcolor{black!6}DualMem & \cellcolor{black!6}\textbf{74.6} & \cellcolor{black!6}[73.8, 75.4] & \cellcolor{black!6}[72.7, 76.6] \\
\midrule
\multirow{5}{*}{MC $50$}
 & Caption        & 58.1 & [56.1, 60.1] & [54.9, 61.9] \\
 & WorldMM-lite   & 27.5 & [25.8, 29.4] & [26.3, 28.7] \\
 & MMA-lite       & 35.2 & [33.4, 37.1] & [32.2, 37.0] \\
 & M2A-lite       & 59.8 & [57.8, 61.7] & [56.2, 63.6] \\
 & \cellcolor{black!6}DualMem & \cellcolor{black!6}\textbf{68.3} & \cellcolor{black!6}[66.4, 70.2] & \cellcolor{black!6}[65.8, 70.6] \\
\bottomrule
\end{tabular}%
}
\caption{\textbf{Cluster-aware 95\% CIs, Qwen2.5-VL-7B.} Naive CIs resample probes iid; cluster CIs resample chains with replacement, the correct unit of independence when probes share an encoding prefix. $10{,}000$ bootstrap resamples.}
\label{tab:cluster-ci-qwen}
\end{table}

\begin{table}[!htb]
\centering
\scriptsize
\setlength{\tabcolsep}{3pt}
\renewcommand{\arraystretch}{1.05}
\resizebox{\columnwidth}{!}{%
\begin{tabular}{@{}l l c c c@{}}
\toprule
$J$ & \textbf{Arch} & \textbf{Mean (\%)} & \textbf{Naive 95\% CI} & \textbf{Cluster 95\% CI} \\
\midrule
\multirow{5}{*}{$5$}
 & Caption        & 58.9 & [57.1, 60.8] & [55.6, 61.9] \\
 & WorldMM-lite   & 43.5 & [41.7, 45.3] & [39.8, 47.2] \\
 & MMA-lite       & 46.1 & [44.3, 48.0] & [42.8, 49.5] \\
 & M2A-lite       & 65.7 & [63.9, 67.5] & [62.5, 68.7] \\
 & \cellcolor{black!6}DualMem & \cellcolor{black!6}\textbf{82.7} & \cellcolor{black!6}[81.3, 84.1] & \cellcolor{black!6}[80.6, 84.5] \\
\midrule
\multirow{5}{*}{$10$}
 & Caption        & 55.6 & [54.3, 57.0] & [53.5, 57.7] \\
 & WorldMM-lite   & 41.1 & [39.7, 42.5] & [38.5, 43.6] \\
 & MMA-lite       & 44.8 & [43.4, 46.2] & [41.8, 47.6] \\
 & M2A-lite       & 65.2 & [63.8, 66.5] & [62.6, 67.7] \\
 & \cellcolor{black!6}DualMem & \cellcolor{black!6}\textbf{75.3} & \cellcolor{black!6}[74.1, 76.4] & \cellcolor{black!6}[72.0, 78.1] \\
\midrule
\multirow{5}{*}{$15$}
 & Caption        & 50.5 & [49.9, 51.2] & [49.5, 51.5] \\
 & WorldMM-lite   & 38.2 & [37.6, 38.9] & [37.0, 39.4] \\
 & MMA-lite       & 35.8 & [35.2, 36.4] & [31.3, 39.4] \\
 & M2A-lite       & 61.9 & [61.3, 62.5] & [60.2, 63.5] \\
 & \cellcolor{black!6}DualMem & \cellcolor{black!6}\textbf{71.3} & \cellcolor{black!6}[70.7, 71.9] & \cellcolor{black!6}[70.3, 72.3] \\
\midrule
\multirow{5}{*}{MC $50$}
 & Caption        & 47.7 & [45.7, 49.7] & [47.4, 48.0] \\
 & WorldMM-lite   & 37.0 & [35.1, 38.9] & [34.7, 39.1] \\
 & MMA-lite       & 36.9 & [35.0, 38.8] & [35.7, 37.9] \\
 & M2A-lite       & 64.7 & [62.8, 66.6] & [63.7, 65.6] \\
 & \cellcolor{black!6}DualMem & \cellcolor{black!6}\textbf{65.1} & \cellcolor{black!6}[63.2, 67.0] & \cellcolor{black!6}[62.8, 67.3] \\
\bottomrule
\end{tabular}%
}
\caption{\textbf{Cluster-aware 95\% CIs, Gemini~2.5~Flash.} Same protocol as Table~\ref{tab:cluster-ci-qwen}. The largest naive-vs-cluster gap is MMA-lite at $J{=}15$, where the cluster interval is roughly seven times wider than the naive one, showing how much the iid assumption can understate variance on the Gemini back-end.}
\label{tab:cluster-ci-gem}
\end{table}

\paragraph{Paired cluster permutation test (DualMem vs M2A-lite).}
Because DualMem and M2A-lite use the same replayed trajectories and common probe IDs, we form the paired outcome difference $d_i = \mathbf{1}[\text{DualMem correct}_i] - \mathbf{1}[\text{M2A-lite correct}_i] \in \{-1,0,+1\}$ for every common probe. For each seeded chain we jointly flip the signs of all $d_i$ values and enumerate all $2^K$ sign assignments for the $K$ chains in the cell which preserves the within-chain dependence and produces exact two-sided $p$-values i.e.  $\Pr(|\bar{d}^{\mathrm{perm}}|\!\geq\!|\bar{d}^{\mathrm{obs}}|)$. Table~\ref{tab:paired-perm} shows the results. All six exhaustive $J\in\{5,10,15\}$ cells have $p \leq 0.0059$. The $J=50$ pilots only contain five chains each, and here Qwen has $+8.52$ points with $p=0.0625$ and Gemini has $+0.45$ points with $p=0.6875$.

\begin{table}[!htb]
\centering
\small
\setlength{\tabcolsep}{5pt}
\renewcommand{\arraystretch}{1.05}
\begin{tabular}{@{}llrrr@{}}
\toprule
\textbf{Back-end} & \textbf{Cell} & $\bar{d}$ \textbf{(pp)} & $p$\textbf{-value} & \textbf{\#chains} \\
\midrule
Qwen   & $J{=}5$       & $+12.22$ & $0.0059$ & 10 \\
Qwen   & $J{=}10$      & $+11.11$ & $0.0015$ & 12 \\
Qwen   & $J{=}15$      & $+11.97$ & $<\!10^{-6}$ & 25 \\
Qwen   & MC $J{=}50$   & $+8.52$ & $0.0625$ & 5 \\
\midrule
Gemini & $J{=}5$       & $+16.98$ & $<\!10^{-6}$ & 25 \\
Gemini & $J{=}10$      & $+10.09$ & $0.0020$ & 10 \\
Gemini & $J{=}15$      & $+9.42$ & $<\!10^{-6}$ & 22 \\
Gemini & MC $J{=}50$   & $+0.45$ & $0.6875$ & 5 \\
\bottomrule
\end{tabular}
\caption{\textbf{Paired cluster permutation test, DualMem (ours) vs M2A-lite (runner-up).} $\bar{d}$ is the mean per-probe outcome difference in percentage points; $p$-values from sign permutations. All six exhaustive cells have $p\leq0.0059$. With five chains, the Qwen $+8.52$-point difference gives $p=0.0625$, the smallest attainable two-sided value; the Gemini $+0.45$-point difference gives $p=0.6875$.}
\label{tab:paired-perm}
\end{table}

\section{More results: per-reach task success rate}
\label{sec:appendix-results}

Tables~\ref{tab:tsr-j5-qwen}--\ref{tab:tsr-mc-qwen} (Qwen2.5-VL-7B) and Tables~\ref{tab:tsr-j5-gem}--\ref{tab:tsr-mc-gem} (Gemini~2.5~Flash) give the full per-reach task success rate (TSR) for all four chain-length settings, one table per $J$ per back-end. Rows are memory architectures; columns are reaches $r$ (number of session boundaries between visit and probe). For the Monte Carlo $J{=}50$ pilot, we bin reaches $r\in[1,49]$ into seven contiguous groups of seven; the underlying per-reach values are sparse (10 probes per reach per chain).

\begin{table}[!htb]
\centering
\tiny
\setlength{\tabcolsep}{4pt}
\renewcommand{\arraystretch}{1.1}
\begin{tabular}{@{}l c ccccc@{}}
\toprule
\textbf{Memory} & \textbf{SR} & $r{=}1$ & $r{=}2$ & $r{=}3$ & $r{=}4$ \\
\midrule
NoMemory & 0.0  & 0.0  & 0.0  & 0.0  & 0.0  \\
TextOnly & 0.0  & 0.0  & 0.0  & 0.0  & 0.0  \\
Caption  & 67.9 & 66.3 & 68.8 & 67.9 & 72.2 \\
WorldMM-lite  & 37.3 & 33.7 & 37.6 & 40.7 & 47.2 \\
MMA-lite      & 47.7 & 48.9 & 46.8 & 42.0 & 58.3 \\
M2A-lite      & 69.0 & 67.4 & 68.8 & 70.4 & 75.0 \\
\rowcolor{black!6}
DualMem \textbf{(ours)} & \textbf{81.2} & \textbf{79.9} & \textbf{83.0} & \textbf{81.5} & \textbf{80.6} \\
\bottomrule
\end{tabular}
\caption{Per-reach TSR (\%) on Qwen2.5-VL-7B, $J{=}5$ ($n_r{=}442$).} 
\label{tab:tsr-j5-qwen}
\label{tab:tsr-all}
\end{table}

\begin{table}[!htb]
\centering
\tiny
\setlength{\tabcolsep}{3pt}
\renewcommand{\arraystretch}{1.1}
\resizebox{\columnwidth}{!}{%
\begin{tabular}{@{}l c ccccccccc@{}}
\toprule
\textbf{Memory} & \textbf{SR} & $r{=}1$ & $r{=}2$ & $r{=}3$ & $r{=}4$ & $r{=}5$ & $r{=}6$ & $r{=}7$ & $r{=}8$ & $r{=}9$ \\
\midrule
NoMemory & 0.0  & 0.0  & 0.0  & 0.0  & 0.0  & 0.0  & 0.0  & 0.0  & 0.0  & 0.0  \\
TextOnly & 0.0  & 0.0  & 0.0  & 0.0  & 0.0  & 0.0  & 0.0  & 0.0  & 0.0  & 0.0  \\
Caption  & 64.5 & 64.7 & 63.5 & 63.9 & 64.7 & 64.5 & 64.3 & 64.4 & 65.7 & 72.9 \\
WorldMM-lite  & 34.1 & 30.7 & 31.6 & 32.1 & 34.2 & 35.9 & 37.2 & 39.6 & 39.0 & 47.9 \\
MMA-lite      & 41.0 & 38.2 & 38.5 & 39.8 & 40.7 & 43.5 & 44.2 & 43.6 & 47.6 & 47.9 \\
M2A-lite      & 66.7 & 67.4 & 66.8 & 67.0 & 68.5 & 65.3 & 63.3 & 65.8 & 65.7 & 70.8 \\
\rowcolor{black!6}
DualMem \textbf{(ours)} & \textbf{77.8} & \textbf{75.4} & \textbf{75.4} & \textbf{78.7} & \textbf{79.0} & \textbf{78.6} & \textbf{78.9} & \textbf{78.5} & \textbf{80.0} & \textbf{87.5} \\
\bottomrule
\end{tabular}%
}
\caption{Per-reach TSR (\%) on Qwen2.5-VL-7B, $J{=}10$ ($n_r{=}2{,}205$).} 
\label{tab:tsr-j10-qwen}
\label{tab:retention-j10-qwen}
\end{table}

\begin{table}[!htb]
\centering
\tiny
\setlength{\tabcolsep}{2pt}
\renewcommand{\arraystretch}{1.1}
\resizebox{\columnwidth}{!}{%
\begin{tabular}{@{}l c cccccccccccccc@{}}
\toprule
\textbf{Memory} & \textbf{SR} & $r{=}1$ & $r{=}2$ & $r{=}3$ & $r{=}4$ & $r{=}5$ & $r{=}6$ & $r{=}7$ & $r{=}8$ & $r{=}9$ & $r{=}10$ & $r{=}11$ & $r{=}12$ & $r{=}13$ & $r{=}14$ \\
\midrule
NoMemory & 0.0  & 0.0  & 0.0  & 0.0  & 0.0  & 0.0  & 0.0  & 0.0  & 0.0  & 0.0  & 0.0  & 0.0  & 0.0  & 0.0  & 0.0  \\
TextOnly & 0.0  & 0.0  & 0.0  & 0.0  & 0.0  & 0.0  & 0.0  & 0.0  & 0.0  & 0.0  & 0.0  & 0.0  & 0.0  & 0.0  & 0.0  \\
Caption  & 62.3 & 64.0 & 63.5 & 62.5 & 62.3 & 62.1 & 60.9 & 61.6 & 61.3 & 61.7 & 62.1 & 61.4 & 61.0 & 62.3 & 72.0 \\
WorldMM-lite  & 29.7 & 30.9 & 29.2 & 28.4 & 28.3 & 28.5 & 28.2 & 29.3 & 29.7 & 31.3 & 32.7 & 30.5 & 31.9 & 33.8 & 38.7 \\
MMA-lite      & 39.4 & 39.6 & 38.2 & 38.7 & 39.1 & 38.6 & 37.6 & 38.4 & 38.7 & 40.8 & 40.8 & 41.4 & 42.4 & 45.4 & 49.5 \\
M2A-lite      & 62.6 & 64.0 & 63.4 & 63.6 & 63.2 & 62.4 & 61.9 & 62.9 & 61.4 & 61.2 & 61.4 & 60.9 & 61.0 & 64.3 & 65.6 \\
\rowcolor{black!6}
DualMem \textbf{(ours)} & \textbf{74.6} & \textbf{75.4} & \textbf{75.7} & \textbf{75.7} & \textbf{75.0} & \textbf{74.7} & \textbf{73.2} & \textbf{73.1} & \textbf{73.1} & \textbf{73.8} & \textbf{75.3} & \textbf{74.0} & \textbf{73.4} & \textbf{76.3} & \textbf{78.5} \\
\bottomrule
\end{tabular}%
}
\caption{Per-reach TSR (\%) on Qwen2.5-VL-7B, $J{=}15$ ($n_r{=}10{,}307$).}
\label{tab:tsr-j15-qwen}
\end{table}

\begin{table}[!htb]
\centering
\tiny
\setlength{\tabcolsep}{3pt}
\renewcommand{\arraystretch}{1.1}
\resizebox{\columnwidth}{!}{%
\begin{tabular}{@{}l c ccccccc@{}}
\toprule
\textbf{Memory} & \textbf{SR} & $r{\in}\![1,7]$ & $r{\in}\![8,14]$ & $r{\in}\![15,21]$ & $r{\in}\![22,28]$ & $r{\in}\![29,35]$ & $r{\in}\![36,42]$ & $r{\in}\![43,49]$ \\
\midrule
NoMemory & 0.0  & 0.0  & 0.0  & 0.0  & 0.0  & 0.0  & 0.0  & 0.0  \\
TextOnly & 0.0  & 0.0  & 0.0  & 0.0  & 0.0  & 0.0  & 0.0  & 0.0  \\
Caption  & 58.1 & 62.6 & 62.6 & 57.4 & 60.6 & 54.9 & 56.0 & 51.8 \\
WorldMM-lite  & 27.5 & 25.7 & 28.9 & 30.3 & 28.0 & 27.4 & 27.7 & 24.4 \\
MMA-lite      & 35.2 & 40.3 & 36.0 & 38.9 & 37.1 & 30.9 & 34.9 & 27.7 \\
M2A-lite      & 59.8 & 68.0 & 65.1 & 62.3 & 61.1 & 54.3 & 56.6 & 49.8 \\
\rowcolor{black!6}
DualMem \textbf{(ours)} & \textbf{68.3} & \textbf{71.7} & \textbf{75.1} & \textbf{68.0} & \textbf{67.1} & \textbf{65.7} & \textbf{66.9} & \textbf{62.9} \\
\bottomrule
\end{tabular}%
}
\caption{Per-reach TSR (\%) on Qwen2.5-VL-7B, Monte Carlo $J{=}50$, reach-binned ($n_r{=}2{,}407$).}
\label{tab:tsr-mc-qwen}
\end{table}

\begin{table}[!htb]
\centering
\tiny
\setlength{\tabcolsep}{4pt}
\renewcommand{\arraystretch}{1.1}
\begin{tabular}{@{}l c ccccc@{}}
\toprule
\textbf{Memory} & \textbf{SR} & $r{=}1$ & $r{=}2$ & $r{=}3$ & $r{=}4$ \\
\midrule
NoMemory & 0.0  & 0.0  & 0.0  & 0.0  & 0.0  \\
TextOnly & 0.0  & 0.0  & 0.0  & 0.0  & 0.0  \\
Caption  & 58.9 & 60.4 & 58.9 & 57.3 & 56.2 \\
WorldMM-lite  & 43.5 & 42.3 & 43.5 & 45.7 & 43.4 \\
MMA-lite      & 46.1 & 47.1 & 46.8 & 44.8 & 43.1 \\
M2A-lite      & 65.7 & 65.4 & 65.7 & 67.4 & 63.7 \\
\rowcolor{black!6}
DualMem \textbf{(ours)} & \textbf{82.7} & \textbf{83.2} & \textbf{82.7} & \textbf{81.5} & \textbf{82.9} \\
\bottomrule
\end{tabular}
\caption{Per-reach TSR (\%) on Gemini 2.5 Flash, $J{=}5$ ($n_r{=}2{,}762$).}
\label{tab:tsr-j5-gem}
\label{tab:tsr-all-gemini}
\end{table}

\begin{table}[!htb]
\centering
\tiny
\setlength{\tabcolsep}{3pt}
\renewcommand{\arraystretch}{1.1}
\resizebox{\columnwidth}{!}{%
\begin{tabular}{@{}l c ccccccccc@{}}
\toprule
\textbf{Memory} & \textbf{SR} & $r{=}1$ & $r{=}2$ & $r{=}3$ & $r{=}4$ & $r{=}5$ & $r{=}6$ & $r{=}7$ & $r{=}8$ & $r{=}9$ \\
\midrule
NoMemory & 0.0  & 0.0  & 0.0  & 0.0  & 0.0  & 0.0  & 0.0  & 0.0  & 0.0  & 0.0  \\
TextOnly & 0.0  & 0.0  & 0.0  & 0.0  & 0.0  & 0.0  & 0.0  & 0.0  & 0.0  & 0.0  \\
Caption  & 55.6 & 56.1 & 56.7 & 55.2 & 54.8 & 55.4 & 53.1 & 56.0 & 56.8 & 58.4 \\
WorldMM-lite  & 41.1 & 40.0 & 41.0 & 40.5 & 39.6 & 40.7 & 41.3 & 44.2 & 46.7 & 45.1 \\
MMA-lite      & 44.8 & 45.1 & 45.6 & 44.6 & 44.1 & 45.0 & 41.7 & 45.1 & 44.5 & 51.3 \\
M2A-lite      & 65.2 & 64.6 & 64.5 & 65.5 & 65.2 & 65.2 & 63.8 & 65.2 & 69.0 & 69.9 \\
\rowcolor{black!6}
DualMem \textbf{(ours)} & \textbf{75.3} & \textbf{74.6} & \textbf{74.8} & \textbf{75.4} & \textbf{75.8} & \textbf{76.2} & \textbf{72.5} & \textbf{74.0} & \textbf{78.6} & \textbf{83.2} \\
\bottomrule
\end{tabular}%
}
\caption{Per-reach TSR (\%) on Gemini 2.5 Flash, $J{=}10$ ($n_r{=}5{,}003$).} 
\label{tab:tsr-j10-gem}
\end{table}

\begin{table}[!htb]
\centering
\tiny
\setlength{\tabcolsep}{2pt}
\renewcommand{\arraystretch}{1.1}
\resizebox{\columnwidth}{!}{%
\begin{tabular}{@{}l c cccccccccccccc@{}}
\toprule
\textbf{Memory} & \textbf{SR} & $r{=}1$ & $r{=}2$ & $r{=}3$ & $r{=}4$ & $r{=}5$ & $r{=}6$ & $r{=}7$ & $r{=}8$ & $r{=}9$ & $r{=}10$ & $r{=}11$ & $r{=}12$ & $r{=}13$ & $r{=}14$ \\
\midrule
NoMemory & 0.0  & 0.0  & 0.0  & 0.0  & 0.0  & 0.0  & 0.0  & 0.0  & 0.0  & 0.0  & 0.0  & 0.0  & 0.0  & 0.0  & 0.0  \\
TextOnly & 0.0  & 0.0  & 0.0  & 0.0  & 0.0  & 0.0  & 0.0  & 0.0  & 0.0  & 0.0  & 0.0  & 0.0  & 0.0  & 0.0  & 0.0  \\
Caption  & 50.5 & 53.5 & 52.1 & 51.4 & 50.5 & 50.1 & 50.0 & 49.1 & 48.9 & 49.0 & 49.4 & 48.9 & 47.8 & 48.8 & 45.1 \\
WorldMM-lite  & 38.2 & 37.9 & 38.2 & 38.3 & 37.9 & 37.6 & 37.8 & 38.7 & 38.5 & 37.9 & 38.1 & 38.0 & 40.4 & 42.4 & 40.2 \\
MMA-lite      & 35.8 & 38.5 & 37.5 & 36.6 & 36.0 & 35.3 & 34.4 & 34.8 & 34.3 & 35.0 & 35.0 & 34.1 & 33.4 & 32.1 & 32.9 \\
M2A-lite      & 61.9 & 62.6 & 62.2 & 62.5 & 62.1 & 62.1 & 62.2 & 61.6 & 61.1 & 61.9 & 61.5 & 60.9 & 59.6 & 60.7 & 56.5 \\
\rowcolor{black!6}
DualMem \textbf{(ours)} & \textbf{71.3} & \textbf{73.6} & \textbf{72.7} & \textbf{72.1} & \textbf{71.5} & \textbf{70.6} & \textbf{70.2} & \textbf{69.3} & \textbf{69.7} & \textbf{69.9} & \textbf{71.2} & \textbf{71.1} & \textbf{69.9} & \textbf{70.2} & \textbf{70.3} \\
\bottomrule
\end{tabular}%
}
\caption{Per-reach TSR (\%) on Gemini 2.5 Flash, $J{=}15$ ($n_r{=}25{,}196$).} 
\label{tab:tsr-j15-gem}
\end{table}

\begin{table}[!htb]
\centering
\tiny
\setlength{\tabcolsep}{3pt}
\renewcommand{\arraystretch}{1.1}
\resizebox{\columnwidth}{!}{%
\begin{tabular}{@{}l c ccccccc@{}}
\toprule
\textbf{Memory} & \textbf{SR} & $r{\in}\![1,7]$ & $r{\in}\![8,14]$ & $r{\in}\![15,21]$ & $r{\in}\![22,28]$ & $r{\in}\![29,35]$ & $r{\in}\![36,42]$ & $r{\in}\![43,49]$ \\
\midrule
NoMemory & 0.0  & 0.0  & 0.0  & 0.0  & 0.0  & 0.0  & 0.0  & 0.0  \\
TextOnly & 0.0  & 0.0  & 0.0  & 0.0  & 0.0  & 0.0  & 0.0  & 0.0  \\
Caption  & 47.7 & 49.4 & 45.1 & 50.6 & 47.1 & 45.4 & 47.4 & 48.7 \\
WorldMM-lite  & 37.0 & 34.9 & 40.3 & 37.1 & 32.6 & 32.3 & 38.3 & 43.5 \\
MMA-lite      & 36.9 & 37.1 & 34.6 & 38.0 & 38.6 & 36.0 & 35.7 & 38.4 \\
M2A-lite      & 64.7 & 64.0 & 65.1 & 67.7 & 62.6 & 61.7 & 65.4 & 66.2 \\
\rowcolor{black!6}
DualMem \textbf{(ours)} & \textbf{65.1} & \textbf{69.4} & \textbf{58.6} & \textbf{66.6} & \textbf{66.9} & \textbf{64.6} & \textbf{63.7} & \textbf{66.2} \\
\bottomrule
\end{tabular}%
}
\caption{Per-reach TSR (\%) on Gemini 2.5 Flash, Monte Carlo $J{=}50$, reach-binned ($n_r{=}2{,}449$).}
\label{tab:tsr-mc-gem}
\end{table}

\end{document}